\documentclass{article}

\usepackage{iftex}
\ifPDFTeX
  \usepackage{CJKutf8}
\else
  \usepackage{xeCJK}
\fi
\usepackage{tabularx}
\usepackage{longcat_style}
\ifPDFTeX\else
  \setCJKmonofont{FandolFang-Regular.otf}
\fi
\usepackage{longtable}

\ifPDFTeX
  \usepackage[utf8]{inputenc}
  \usepackage[T1]{fontenc}
\fi
\usepackage{hyperref}
\usepackage{xcolor}
\usepackage{float}
\hypersetup{
    colorlinks=true,
    linkcolor=blue,
    urlcolor=blue,
    citecolor=blue,
    linkbordercolor=blue,
    urlbordercolor=blue,
    citebordercolor=blue,
    pdfborderstyle={/S/U/W 1},
}

\usepackage{xurl}
\usepackage{booktabs}
\usepackage{amsfonts}
\usepackage{nicefrac}
\usepackage{microtype}
\usepackage{graphicx}
\usepackage{natbib}
\usepackage{doi}
\usepackage{amsmath}
\usepackage{amssymb}
\usepackage{xspace}
\usepackage{enumitem}
\usepackage{needspace}
\usepackage{multirow}
\usepackage{subcaption}
\usepackage{makecell}

\usepackage{hyperref, cleveref}
\setlist[itemize]{leftmargin=*}
\setlist[enumerate]{leftmargin=*}
\setlist[description]{leftmargin=*}


\usepackage{colortbl}
\definecolor{mygray}{gray}{.88}
\definecolor{mycyan}{cmyk}{.15,0,0,0}
\definecolor{mycyan2}{cmyk}{.85,0,0,0}
\definecolor{mygreen}{rgb}{0.19,0.79,0.02}
\definecolor{midnightgreen}{rgb}{0.0,0.29,0.33}

\title{AutoRL: A Multi-Agent System towards Autonomous Reinforcement Learning}
\newcommand{\authoraffil}[2]{\textbf{#1\textsuperscript{#2}}}
\author{%
  \authoraffil{Simon Tang}{1,2}, \authoraffil{Wenjie Shi}{1,*}, \authoraffil{Yongjing Yin}{1}, \authoraffil{Keheng Wang}{1}, \authoraffil{Jialin Liu}{1}, \\ \authoraffil{Zheng Zhang}{1,3}, \authoraffil{Linvo Lin}{4}, \authoraffil{Xiangyuan Xue}{2}, \authoraffil{Kan Ren}{3}, \authoraffil{Jingang Wang}{1}, \authoraffil{Xunliang Cai}{1} \\
  \textsuperscript{1}Meituan \quad \textsuperscript{2}The Chinese University of Hong Kong \quad \textsuperscript{3}ShanghaiTech University \quad \textsuperscript{4}Fudan University\\
  \textsuperscript{*}Corresponding author \\
  shiwenjie05@meituan.com
}
\date{}

\renewcommand{\headeright}{\raisebox{-0.2\height}{\includegraphics[height=1.8em]{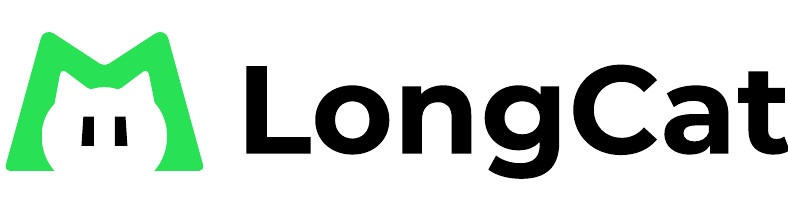}}}
\renewcommand{\shorttitle}{AutoRL}
\hypersetup{
  pdftitle={AutoRL: A Multi-Agent System towards Autonomous Reinforcement Learning},
  pdfauthor={Simon Tang, Wenjie Shi, Yongjing Yin, Keheng Wang, Jialin Liu, Zheng Zhang, Linvo Lin, Xiangyuan Xue, Kan Ren, Jingang Wang, Xunliang Cai},
  pdfsubject={Automating reinforcement learning supervision and research},
  pdfkeywords={AutoRL, autonomous reinforcement learning, multi-agent systems}
}

\begin{document}
\ifPDFTeX\begin{CJK*}{UTF8}{gbsn}\fi
\maketitle

\begin{abstract}
Recursive self-improvement (RSI) is one of the most promising paths toward an AI singularity for superintelligence. Achieving this vision requires AI systems to undertake the full training pipeline for iterative improvement. Reinforcement learning (RL) plays an essential role in this cycle to improve model capabilities and generalization, yet automating RL remains difficult due to unstable training dynamics and the need for sustained supervision. We introduce \textbf{AutoRL}, a multi-agent system for \textbf{automating RL training}, as a cornerstone for future RSI. Different from conducting research by end-to-end complete experiments, AutoRL monitors the evolving training state and integrates \textbf{process-level supervision and research into an ongoing RL training}, following the practical workflow of an RL researcher. Specifically, AutoRL consists of three RL-specific agentic roles. Perceptors track training dynamics, rollouts, evaluation, and efficiency, triggering deliberation when they detect anomalies, degradation, or stagnation. Three complementary Proposers diagnose problems and refine candidate interventions through shared evidence and checks. The Orchestrator reviews their recommendation and applies approved changes through online hot updates or checkpoint-based resumption.
Experiments on Qwen3 and LongCat-Flash-Lite demonstrate the superiority of AutoRL over static RL recipes and a general agent. On Qwen3-4B, AutoRL outperforms both baselines across nine benchmarks. In the coding-agent experiment, automated supervision supports 14.7 days of training with 11 runtime failure fixes, with pass@1 increasing 12.84\% on SWE-related tasks.
The research records show that existing agents can handle unattended RL supervision and dynamic adjustment, such as hyperparameter tuning, implementation repair, and reward refinement. The difficulty for future RSI lies in developing fundamental methods that are valuable for resource-consuming validation. 
\end{abstract}
\begin{figure}[H]
  \centering
  \includegraphics[width=\linewidth]{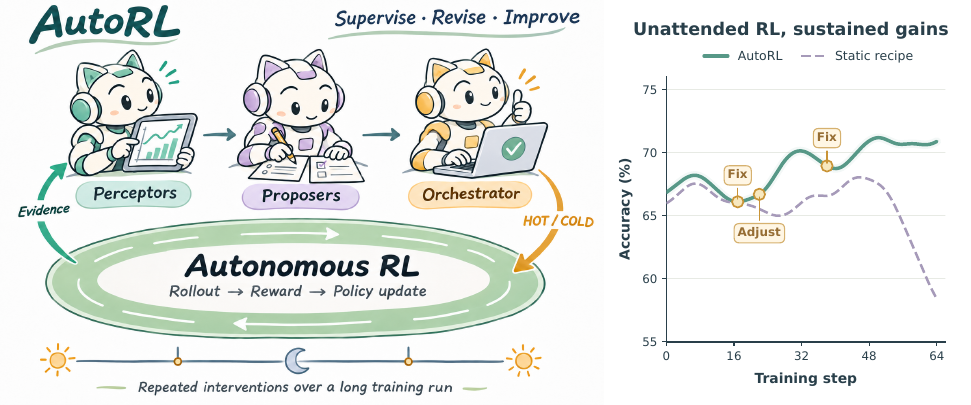}
  \caption{\textbf{AutoRL: autonomous RL supervision and research.} Right: AutoRL continually adopts interventions, such as fixes and adjustments, to sustain gains over a static recipe.}
  \label{fig:autorl-teaser}
\end{figure}

\clearpage
\section{Introduction}
\label{sec:introduction}

Recursive self-improvement (RSI) aims to drive AI systems to iteratively optimize and evolve themselves~\citep{chen2026rsisurvey,zhang2025darwingodel}. As these systems become stronger at developing their successors, RSI could promote an improvement flywheel for intelligence explosion~\citep{chan2026intelligenceexplosion} and an AI singularity~\citep{jafari2025aisingularity}, pointing out a promising route to superintelligence~\citep{zhou2026_261021055}. Sustaining such self-directed improvement remains difficult, as identifying useful changes and reliably assessing their effects still pose substantial challenges~\citep{chen2026rsisurvey}. A practical step pursued by recent work is to place a more powerful existing model in charge of optimizing another model, delegating decisions about training and subsequent experiments to AI agents~\citep{shi2026_260620657,du2026_260935561,zhou2026_261021055}. Within this model-driven optimization process, reinforcement learning (RL) is an essential technique for improving model capabilities and a particularly demanding stage to automate. Learning from reward feedback supports improvements that generalize beyond the training examples~\citep{ouyang2022instructgpt,shao2024deepseekmath}, making RL central to frontier post-training recipes~\citep{kimiteam2026k3,deepseekai2026v41}. Yet maintaining a productive RL run requires responding to a training process with continually dynamic states. As the policy is optimized, the distribution of generated experience evolves, which further alters the learning signal used for later updates. Distributed execution adds numerical and operational failures that can interrupt this process~\citep{fu2025_250524298}. Even a run that executes successfully may stagnate under a flawed reward or an unsuitable recipe. Thus, automating RL requires following the evolving training state and deciding when and how to revise the current procedure.

Existing autonomous-training systems largely organize research across successive experiments, using the results of one attempt to inform the next recipe~\citep{fang2026llmzero,chen2026evotrainer,he2026autosft,zhou2026_261021055}. Differently, human RL researchers also investigate the experiment while it is running. A change in reward may lead them to inspect generated responses and repair the grader, while a persistent plateau can prompt reconsideration of the recipe. They apply supported adjustments directly or resume from a checkpoint when a modification requires restarting, then examine the subsequent behavior. Motivated by human RL research workflow, we focus on \textbf{integrating this process-level supervision and research into an ongoing RL experiment}. Within a predefined task and resource budget, the agent system uses the evolving training state to guide investigation and execution. There are three practical challenges for process-level autonomous RL.

\begin{itemize}[topsep=4pt,itemsep=5pt,parsep=0pt]
  \item \textbf{RL training requires timely intervention.} Failures, numerical instability, and performance degradation can emerge while training is still running. Waiting until an experiment finishes delays diagnosis and can waste substantial computation. Early stopping and live modification provide useful intervention mechanisms~\citep{fang2026llmzero,fu2026_260604484}, but sustained supervision also requires deciding when to intervene, how to respond, and whether the response worked as training continues.
  \item \textbf{Monitoring and improvement research operate at different rhythms.} Monitoring requires frequent checks and prompt responses, whereas analyzing a plateau and developing a new training strategy require sustained investigation. Handling both in one agent context makes them compete for attention, with execution output potentially crowding out analysis~\citep{he2026autosft}. A long-running system needs to keep watching training while investigating how to improve it.
  \item \textbf{Validating RL hypotheses is expensive.} A plausible proposal may rest on an incorrect diagnosis or contain an implementation error, yet testing every proposal in a separate RL run consumes substantial GPU resources. Independent review can improve proposal scrutiny~\citep{du2026_260935561}, but useful validation also needs to distinguish questions that code inspection and existing evidence can resolve from those that require further training. Effective validation requires making this distinction before committing resources, then assessing the intervention through subsequent feedback.
\end{itemize}

To address these challenges, we propose \textbf{AutoRL}, which organizes the responsibilities of an expert RL researcher into specialized agent roles as shown in Figure~\ref{fig:autorl-overview}. Four \textbf{Perceptors} maintain complementary views of the running experiment. They use generated trajectories and evaluation results to interpret optimization behavior, while tracking execution health. Signals of an anomaly or stalled progress initiate research by three \textbf{Proposers}. Each first develops an independent diagnosis before reviewing the others' proposals. The Tuner explores adjustments to the current recipe, while the Innovator investigates alternative mechanisms. The Reviewer examines implementation correctness and challenges unsupported assumptions. During deliberation, they challenge one another's causal explanations using training records and code evidence, performing CPU-only checks to resolve concrete uncertainties. Proposers then revise, merge, or withdraw candidates in response to these findings before selecting a recommendation. An \textbf{Orchestrator} then decides whether and how to apply the recommendation. It reviews the proposed modification against the current state and applies approved modifications through a HOT online update when supported. Otherwise, it coordinates a COLD resumption from a compatible checkpoint, which also provides a path for repairing and recovering a failed run. Confirming runtime activation establishes which recipe governs subsequent observations. RL-specific tools support these operations, while persistent memory and shared records retain the evidence behind earlier decisions. Perceptors continue observing the resulting training behavior and trigger further investigation when needed for the next loop. This division keeps monitoring active during research and carries research conclusions into training as it progresses.

We evaluate AutoRL on mathematical reasoning and agentic coding tasks with Qwen3~\citep{yang2025qwen3} and LongCat-Flash-Lite~\citep{liu2026longcatflashlite}. On Qwen3-4B, AutoRL yields stronger trained models than static RL recipes and supervision by the native general agent. The advantage holds across all nine benchmarks, extending beyond mathematics to general knowledge and coding. Training records show how repairing a flawed recipe and investigating a plateau lead to renewed gains. During the 14.7-day coding-agent run, checkpoint-based recovery allows training to continue after 11 runtime failures. On SWE-related tasks, pass@1 rises from 52.29\% to 65.13\%, demonstrating AutoRL can obtain continued improvement on unattended long-term RL experiments across interruptions. Compared with the general agent, AutoRL achieves these gains while investigating more broadly and reverting fewer interventions. Preserving more training progress when applying or withdrawing a modification also allows subsequent research to build on the model's accumulated learning. Meanwhile, in the GPT-driven Qwen3-4B run, 131 initial candidates yield 20 deployed plans, all refining the existing recipe. Three candidates attempt to redesign the learning rule, yet none reaches deployment. It suggests the scope and limitations of current agents in AutoRL: they can monitor the training process and its evolving state, conducting local refinements and dynamically adjusting the recipe, but struggle to produce fundamental methodological breakthroughs in our experiments. We believe that the proposed AutoRL is not the only framework or organizing form for autonomous RL, but we hope it can become a pioneering agentic behavior research for developing future RSI.

\Needspace{4\baselineskip}
Our contributions are as follows:
\begin{itemize}
  \item We present a pioneering study of process-level autonomous RL as a crucial component of future RSI, analyzing the challenges of sustaining supervision and research within an ongoing RL experiment.
  \item We introduce AutoRL, a multi-agent framework in which Perceptors monitor training and signal the need for research. Proposers discuss diagnoses and refine interventions using shared evidence. The Orchestrator reviews and applies their recommendations, while Perceptors observe the effects to inform subsequent research and adjustment.
  \item Experiments show that AutoRL sustains unattended RL monitoring and research, delivering continued gains over static recipes and a general agent. Behavioral analysis reveals that current agents can monitor training and dynamically adjust recipes, but are limited in delivering fundamental methodological breakthroughs in our experiments.
\end{itemize}

\section{AutoRL}
\label{sec:method}
\begin{figure}[t]
  \centering
  \includegraphics[width=\linewidth]{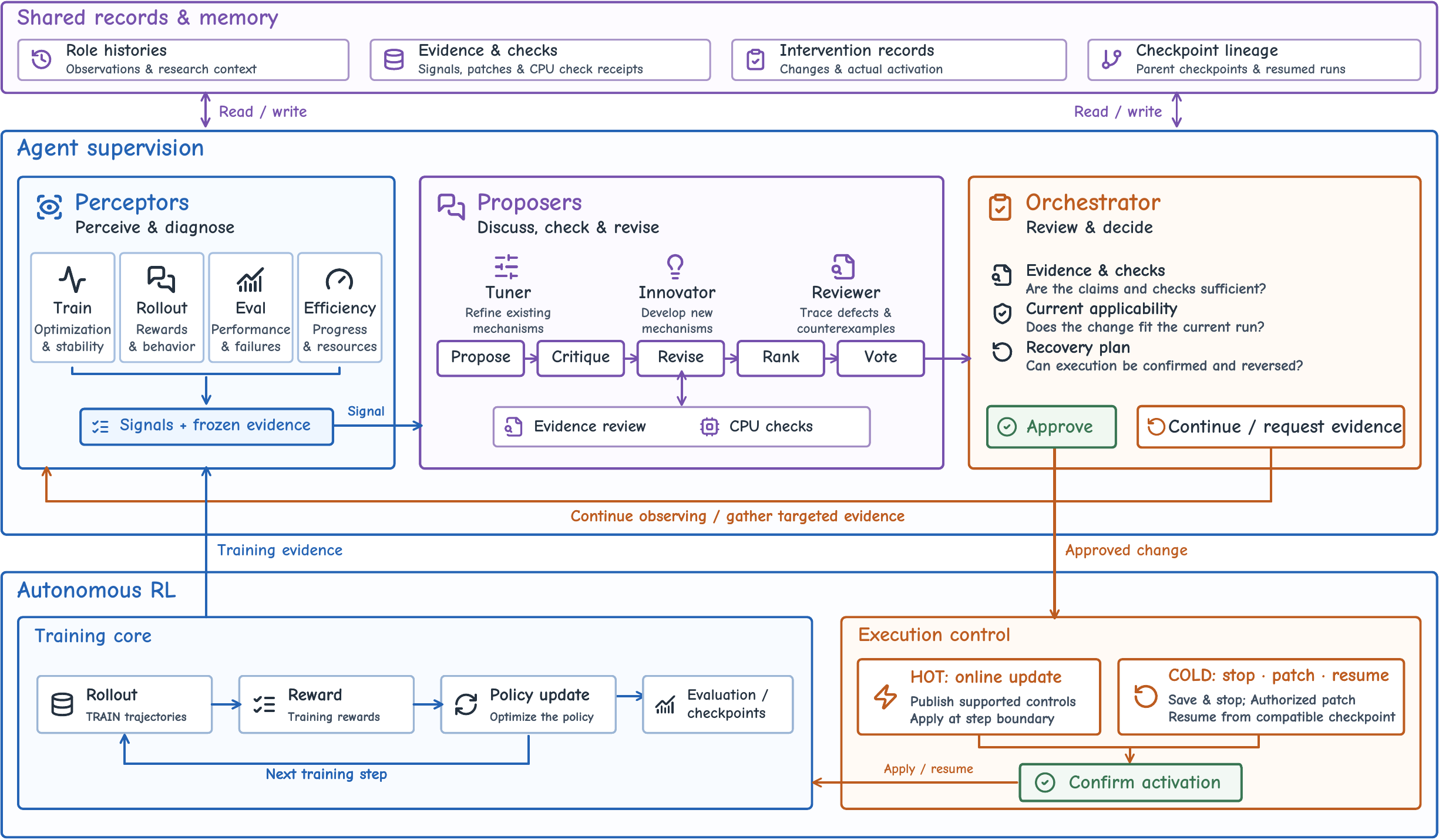}
  \caption{\textbf{Framework of AutoRL.} Four Perceptors monitor training and diagnose problems. Three specialized Proposers discuss candidates, review evidence, and perform CPU-only checks to guide revision and selection. The Orchestrator reviews the recommendation, authorizes a HOT update or COLD checkpoint-based resumption, and confirms runtime activation. Shared records retain evidence, interventions, and checkpoint lineage.}
  \label{fig:autorl-overview}
\end{figure}

\subsection{Task and Monitoring Loop}
\label{sec:method-overview}
Given an initial model, training data, objectives, a resource budget, and authorized tools specified by the researcher, AutoRL automates training supervision and investigates improvements within this task. It aims to reduce the need for continuous human monitoring and intervention while improving the model under a fixed evaluation protocol. As shown in Figure~\ref{fig:autorl-overview}, it couples RL training with agent-based supervision. The training process repeatedly performs rollout generation, reward computation, and policy updates, with periodic evaluation and checkpointing. Separate supervisory roles handle monitoring, research, and execution control alongside the training process. After the Orchestrator launches the main training run, four Perceptors monitor optimization, rollouts, evaluation, and efficiency. Signals of anomalies, degradation, or stagnation trigger three Proposers to independently diagnose the issue, challenge candidate interventions, and refine their proposals. The Orchestrator evaluates their recommendation, applies approved changes through an online hot update or checkpoint-based cold resumption, and confirms runtime activation. Intervention records inform subsequent perception and deliberation, which continue alongside training. Stopping or resumption occurs when needed for an approved change or failure recovery.

Research decisions adapt the active training run using current evidence and prior interventions. We denote the $k$-th decision as
\begin{equation}
  a_k = \mathcal F(E_k,H_k),
  \label{eq:research-procedure}
\end{equation}
where $E_k$ contains the training observations and evidence used in the decision, $H_k$ denotes retrievable training and intervention history, and $a_k$ is the action taken after the Orchestrator's review. The procedure $\mathcal F$ is implemented by the specialized roles, evidence-based deliberation, and execution controls described below. The index $k$ counts research decisions, while $(r,s)$ identifies a training position by its run $r$ and discrete step $s$.

\subsection{Specialized Roles and Shared Memory}
\label{sec:method-perception}
To alleviate context pressure and concentrate attention on relevant evidence, AutoRL separates the training lifecycle into management, continuous perception, and improvement research. Each role operates in its own context and coordinates through shared evidence and run records:
\begin{itemize}
  \item \textbf{Orchestrator: coordination and execution control.} The Orchestrator manages launch, stopping, resumption, and rollback, schedules observations and deliberation, and evaluates recommendations against the current training state. It coordinates approved interventions, verifies their actual activation, and records the resulting intervention and recovery history.
  \item \textbf{Perceptors: continuous perception and diagnosis.} Four Perceptors examine complementary training signals. The Train Perceptor monitors gradients, losses, and numerical stability; the Rollout Perceptor examines training rewards, sample quality, and generated behavior; the Eval Perceptor tracks performance and failure modes under the fixed evaluation protocol; and the Efficiency Perceptor monitors progress, throughput, and runtime failures. RL-specific tools and procedures support each role's analysis, linking issue signals to the relevant run, step, and evidence window.
  \item \textbf{Proposers: improvement research and candidate review.} Three Proposers independently examine the same evidence. The Tuner develops local adjustments to existing mechanisms; the Innovator explores alternative mechanisms within the task; and the Reviewer traces implementation defects and counterexamples. Cross-examination, code inspection, and applicable CPU-only checks refine their candidates into a recommendation with a concrete modification, expected effects, and supporting evidence.
\end{itemize}

Specialized perception turns training observations into actionable research questions. Each Perceptor follows a predefined workflow for collecting fresh evidence, using tools to compute rolling statistics and trends. These procedures guide diagnosis, while the LLM interprets the evidence and chooses focused follow-up queries. For example, it checks a gradient spike against losses and clipping, or examines falling entropy in the context of training progress to distinguish collapse from ordinary convergence. Explicit rules cover clear triggers, including verified NaN/Inf values and a configured streak of evaluations without a new best score. Before issuing a signal, a Perceptor links the observed behavior to its run, step, measurement window, and source artifacts. A synchronous observation barrier gathers the relevant reports and freezes their signals and initial evidence into a common discussion batch. An ordinary observation cycle without a signal leaves training to continue.

When a decision depends on a missing fact, the Orchestrator can request a targeted observation from the appropriate role. Persistent memory connects these decisions across the training lifecycle. Local histories retain role-specific observations and experience, while shared records connect runs, evidence, candidates, checks, and interventions. Retrieving these records allows agents to reuse prior analysis and check its relevance to the current state. This division of responsibilities aligns each role's context, evidence access, and execution authority.

\subsection{Evidence-based Deliberation}
\label{sec:method-deliberation}
AutoRL develops intervention proposals by combining independent diagnosis with evidence-based deliberation. The three Proposers start from the same initial evidence and examine it from their complementary perspectives. Each candidate specifies the target problem, a mechanism hypothesis, a concrete modification, the expected observable effects, and unresolved questions. Independent initial judgments bring distinct directions into the discussion, where cross-examination tests their premises. Deliberation proceeds through proposal generation, critique, revision, ranking, and voting, for at most five rounds.

During proposal development and revision, Proposers inspect code, metrics, and historical evidence. They can perform applicable CPU-only tests or replay stored training trajectories in isolated working directories. For example, a reward-rule change can be checked against past misjudgments and previously correct cases, while a filtering change can be checked for sample-group integrity and replenishment behavior. Results are shared between rounds and support candidate revision, withdrawal, or merging. The resulting evidence provides the basis for ranking the surviving candidates. Each Proposer ranks candidates by their expected benefit under the fixed evaluation protocol and the strength of the supporting evidence, considering whether intervention is needed. Risk and reversibility also matter, together with the cost of interrupting training and any overlap with other candidates.

The protocol aggregates these individual rankings using Borda scoring. With $m$ surviving candidates, each ranking awards $m$ points to the first choice, decreasing by one to a single point for the last. Scores are summed across the three Proposers, and at most two candidates advance, with ties resolved by candidate identifier. Each Proposer then casts one vote with a rationale, producing a 3:0 or 2:1 recommendation. A sole finalist still requires three confirmation votes. Deliberation can end earlier when all three Proposers explicitly endorse the same eligible candidate in a completed round. In either case, failed checks or unresolved blockers prevent publication. The recommendation retains dissenting views and unresolved questions.

The progression from independent proposals to a final recommendation is
\begin{equation}
  \begin{aligned}
    \mathcal C_k^0 &= \bigcup_{j=1}^{3}\mathcal G_j(E_k^0,H_k),\\
    \widehat a_k &= \operatorname{Debate}(\mathcal C_k^0,E_k,H_k),
  \end{aligned}
  \label{eq:proposal-deliberation}
\end{equation}
where $E_k^0$ is the frozen initial evidence, $\mathcal G_j$ is the proposal procedure of Proposer $j$, and $\mathcal C_k^0$ collects their initial candidates. Evidence gathering and CPU-only tests augment $E_k$ and inform candidate revision and selection. The Orchestrator reviews $\widehat a_k$ and its evidence to determine $a_k$, which may apply a modification or leave the recipe unchanged. Inexpensive checks address logic and implementation issues that can be tested in advance, while subsequent training feedback assesses the remaining questions. This integrates candidate development and validation within deliberation and directs GPU training resources toward approved interventions in the main run.

\subsection{Process-Level Intervention}
\label{sec:method-intervention}
The Orchestrator reviews the evidence supporting a recommendation and approves, defers, rejects, or requests further investigation. Before approving a change, it checks that the expected effect can be assessed against a pre-intervention reference and identifies the evidence needed for the next review. Interventions can change rewards, losses, advantage computation, or data selection, and can repair implementation defects. The task configuration defines the permitted scope, while the evaluation protocol remains fixed. Researchers retain control over changes to the task objectives, access to additional data sources, and the budget for subsequent stages. Actions on the main training process take three forms:
\begin{equation}
  a_k \in \bigl\{\mathrm{NONE},\;\mathrm{HOT}(\delta_k),\;
                   \mathrm{COLD}(\delta_k,b_k)\bigr\}.
  \label{eq:action-space}
\end{equation}
Here, $\delta_k$ is a control update, code patch, or data modification, and $b_k$ is a compatible, complete checkpoint. NONE leaves the training configuration unchanged. HOT applies controls supported by the running backend at an allowed training-step boundary. A Proposer selects this channel only when the complete modification fits the documented online-control interface and is permitted by the task configuration. The Orchestrator verifies that the running backend actually consumes the proposed controls. COLD handles code, configuration, or data changes that require stopping the run. For example, replacing a supported policy-loss object can be HOT, whereas changing the data loader or editing training code requires COLD. The readiness check enforces this distinction by rejecting HOT patches that modify files outside the online-control file.

For a COLD intervention, the Orchestrator coordinates checkpointing and shutdown, authorizes the patch and regression checks, and resumes from a compatible checkpoint. When the run has already failed, recovery uses an existing complete checkpoint. These paths allow adjustments and repairs to preserve accumulated training progress.

To express how the two intervention paths preserve training progress, let $x_k^-=(\theta_k^-,z_k^-,c_k^-)$ denote the state immediately before execution, including model parameters, training runtime state, and effective configuration. The intervention changes this state according to
\begin{equation}
  x_k^+=
  \begin{cases}
    x_k^-, & a_k=\mathrm{NONE},\\
    (\theta_k^-,z_k^-,c_k^-\oplus\delta_k),
      & a_k=\mathrm{HOT}(\delta_k),\\
    (\theta_{b_k},z_{b_k},c_k^-\oplus\delta_k),
      & a_k=\mathrm{COLD}(\delta_k,b_k).
  \end{cases}
  \label{eq:intervention-state}
\end{equation}
Here, $z$ groups the optimizer, trainer, rollout, and data-consumption states. The operator $\oplus$ applies the modification and reconciles which effective controls carry forward. HOT retains the current parameters and runtime state, while COLD restores them from $b_k$ and continues with the modified configuration.

Training then continues from the resulting state. The Orchestrator records the actual change, its effective position $(r,s)$, intended effects, and recovery information. Checkpoint lineage distinguishes the trajectory inherited by the current run from other attempts retained as research experience. Runtime receipts confirm activation, after which Perceptors use the intervention records to interpret fresh observations. At subsequent reviews, the Orchestrator compares this evidence with the pre-intervention reference, accounting for the effective step and the amount of relevant training completed. For a numerical repair, it checks for finite updates and resumed progress. For a reward adjustment, it examines the behavior of affected training examples, while model performance is assessed through the unchanged evaluation protocol.

Verified fixes and beneficial changes are retained. When relevant samples or evaluations are still pending, the Orchestrator specifies the next observation and continues monitoring. Persistent failure or regression prompts rollback or renewed deliberation to revise the intervention; a new anomaly or plateau can likewise trigger further research through the Perceptors. Follow-up uses the existing observation cycles, with the intervention and its observed effects entering the shared history for subsequent decisions.

\section{Experiments}
\label{sec:experiments}

We conduct RL experiments on mathematical reasoning and agentic coding to evaluate whether AutoRL can automate training supervision and recipe improvement. We first compare the trained models across nine benchmarks to assess performance gains at different model scales. We then study what AutoRL can carry out and how AutoRL achieves these gains during training, focusing on repairing a flawed recipe and overcoming a performance plateau. Finally, we examine a two-week coding-agent run to understand how AutoRL handles recurring failures while preserving learning progress. Together, the experiments show that AutoRL can carry out most tasks in ongoing RL supervision and research, such as recipe repair, plateau investigation, and failure recovery during training.

\subsection{Settings}

\textbf{Models and datasets.} We study four settings. Qwen3-1.7B and Qwen3-4B~\citep{yang2025qwen3} are trained on DAPO-Math-17K~\citep{yu2025dapo}, using 17,390 processed mathematical problems. LongCat-Flash-Lite~\citep{liu2026longcatflashlite} is trained on private collections of 9,340 mathematical problems for reasoning and 700+ repository-level software engineering tasks for agentic coding. Coding-agent evaluation uses SWE-related tasks. We evaluate the Qwen models on AIME 2024/2025~\citep{maa2025aime}, HMMT 2025~\citep{matharena2025hmmt}, Beyond AIME~\citep{seed2025beyondaime}, and IMO-AnswerBench~\citep{luong2025imobench} for mathematics; MMLU-Pro~\citep{wang2024mmlupro} and GPQA-Diamond~\citep{rein2023gpqa} for general knowledge and reasoning; and MBPP~\citep{austin2021mbpp} and LiveCodeBench v6 Hard~\citep{jain2024livecodebench} for coding. AutoRL's supervisory agents use \texttt{gpt-6.1-sol}~\citep{openai2026gpt61sol} in both Qwen experiments and \texttt{glm-5.2}~\citep{zai2026glm52} in both LongCat-Flash-Lite experiments. The general agent baseline uses native Codex~\citep{openai2026codexcli} with all built-in tools and features enabled, backed by the same \texttt{gpt-6.1-sol} model as the Qwen AutoRL runs. Section~\ref{sec:analysis-backbones} additionally evaluates GLM-5.2 as the supervisory backbone for Qwen3-4B.

\textbf{Training recipes.} Both Qwen models start from a GRPO recipe~\citep{shao2024deepseekmath} with a learning rate of $10^{-6}$, sampling eight responses per prompt and allowing up to 8,192 response tokens. LongCat reasoning uses the same learning rate and group size with a 131,072-token response limit. Coding-agent training uses fully asynchronous updates, 16 trajectories per task, and a learning rate of $5\times10^{-6}$, allowing up to 400 interaction turns within a 262,144-token context. AutoRL adapts these initial recipes during training while keeping evaluation protocols fixed. Evaluation uses temperature 0.6 for Qwen, with thinking mode disabled, and 1.0 for LongCat. We sample each question or task four times to compute the final results. We report mean accuracy for reasoning and knowledge tasks and pass@1 for coding. For Qwen comparisons, a training step denotes one rollout-and-training iteration, while cumulative counts include steps repeated after checkpoint rollback. Additional details are provided in Appendices~\ref{app:configurations} and~\ref{app:runtime-control}.

\subsection{Main Results}

To evaluate whether sustained supervision improves the resulting model, we compare AutoRL with static-recipe GRPO and the general agent on Qwen3-1.7B and Qwen3-4B, also reporting a baseline reference. Mathematics benchmarks measure gains on the training domain, while general-knowledge and coding benchmarks assess transfer and generalization ability. Specifically, Tables~\ref{tab:main-results} and~\ref{tab:cross-domain-results} report the main performance comparison. Section~\ref{sec:analysis-general-agent} additionally compares training trajectories over the same number of completed steps.

\begin{table}[!htbp]
\centering
\caption{Mathematics benchmark results on Qwen3 models (\%). The numbers in parentheses show the performance comparison between AutoRL and GRPO (Static Recipe).}
\label{tab:main-results}
\setlength{\tabcolsep}{4pt}
\renewcommand{\arraystretch}{1.15}
\resizebox{\linewidth}{!}{%
\begin{tabular}{llccccc}
\toprule
Model & Method & AIME 24 & AIME 25 & HMMT 25 & Beyond AIME & IMO-AnswerBench \\
\midrule
\multirow{4}{*}{Qwen3-1.7B} & Baseline & 15.00 & 13.33 & 5.00 & 4.50 & 10.25 \\
 & GRPO (Static Recipe) & 26.67 & 21.67 & 18.33 & 11.00 & 8.63 \\
 & General Agent & 31.67 & 18.33 & 11.67 & 12.00 & 15.00 \\
 & AutoRL & 30.00(\textcolor{red}{+3.33}) & 30.00(\textcolor{red}{+8.33}) & 18.33(0.00) & 12.50(\textcolor{red}{+1.50}) & 14.63(\textcolor{red}{+6.00}) \\
\midrule
\multirow{4}{*}{Qwen3-4B} & Baseline & 25.00 & 23.33 & 11.67 & 10.5 & 18.44 \\
 & GRPO (Static Recipe) & 46.67 & 45.00 & 20.00 & 26.00 & 27.00 \\
 & General Agent & 50.00 & 43.33 & 26.67 & 25.00 & 16.06 \\
 & AutoRL & 66.67(\textcolor{red}{+20.00}) & 56.67(\textcolor{red}{+11.67}) & 31.67(\textcolor{red}{+11.67}) & 32.50(\textcolor{red}{+6.50}) & 36.75(\textcolor{red}{+9.75}) \\
\bottomrule
\end{tabular}%
}
\par\vspace{2pt}
\end{table}

\begin{table}[!htbp]
\centering
\caption{General-knowledge and coding results on Qwen3 models (\%). The numbers in parentheses show the performance comparison between AutoRL and GRPO (Static Recipe).}
\label{tab:cross-domain-results}
\setlength{\tabcolsep}{5pt}
\renewcommand{\arraystretch}{1.15}
\resizebox{\linewidth}{!}{%
\begin{tabular}{llcccc}
\toprule
\multicolumn{1}{l}{\multirow{2}{*}[-0.65ex]{Model}} & \multicolumn{1}{l}{\multirow{2}{*}[-0.65ex]{Method}} & \multicolumn{2}{c}{General knowledge} & \multicolumn{2}{c}{Coding} \\
\cmidrule(lr){3-4}\cmidrule(lr){5-6}
 & & MMLU-Pro & GPQA-Diamond & MBPP & LiveCodeBench v6 Hard \\
\midrule
\multirow{4}{*}{Qwen3-1.7B} & Baseline & 38.85 & 28.79 & 33.20 & 1.57 \\
 & GRPO (Static Recipe) & 34.05 & 22.60 & 10.95 & 1.50 \\
 & General Agent & 25.55 & 32.20 & 32.00 & 1.07 \\
 & AutoRL & 44.88(\textcolor{red}{+10.83}) & 34.34(\textcolor{red}{+11.74}) & 40.05(\textcolor{red}{+29.10}) & 2.00(\textcolor{red}{+0.50}) \\
\midrule
\multirow{4}{*}{Qwen3-4B} & Baseline & 55.88 & 43.43 & 62.55 & 4.93 \\
 & GRPO (Static Recipe) & 57.93 & 44.70 & 63.55 & 5.71 \\
 & General Agent & 37.73 & 45.33 & 31.80 & 4.50 \\
 & AutoRL & 59.73(\textcolor{red}{+1.80}) & 47.22(\textcolor{red}{+2.52}) & 65.25(\textcolor{red}{+1.70}) & 6.79(\textcolor{red}{+1.08}) \\
\bottomrule
\end{tabular}%
}
\end{table}

\textbf{Performance across benchmarks.} As shown in Tables~\ref{tab:main-results} and~\ref{tab:cross-domain-results}, AutoRL outperforms static-recipe GRPO on all nine Qwen3-4B benchmarks. The largest mathematics gain is on AIME 2024, where accuracy rises from 46.67\% to 66.67\%, an improvement of 20.00\%. AIME 2025 and HMMT 2025 each improve by 11.67\%, reaching 56.67\% and 31.67\%. Beyond AIME and IMO-AnswerBench improve by 6.50\% and 9.75\%. The results demonstrate that AutoRL can obtain consistent improvement compared with the static RL recipe. Meanwhile, gains also extend to general knowledge and coding, with improvements of 1.80\%, 2.52\%, 1.70\%, and 1.08\% on MMLU-Pro, GPQA-Diamond, MBPP, and LiveCodeBench v6 Hard, respectively. It can be observed that improving the mathematics training recipe benefits tasks outside the training domain in this experiment, suggesting that AutoRL achieves gains on a given task by improving the model’s generalization ability rather than by overfitting to the task.

\textbf{Performance across model scales.} On Qwen3-1.7B, AutoRL reaches 30.00\% on both AIME benchmarks, improving on static-recipe GRPO by 3.33\% on AIME 2024 and 8.33\% on AIME 2025. Beyond AIME improves from 11.00\% to 12.50\%, and IMO-AnswerBench from 8.63\% to 14.63\%, while HMMT 2025 remains at 18.33\%. AutoRL also leads the static recipe on all four general-knowledge and coding benchmarks. These results show that the same supervisory workflow is size-agnostic and can improve both model sizes using feedback from their respective training runs.

\textbf{Comparison with general agent.} AutoRL outperforms the Codex-supervised Qwen3-4B model on all nine benchmarks. AIME 2024 rises from 50.00\% to 66.67\%, and IMO-AnswerBench from 16.06\% to 36.75\%. The gains reach 22.00\% on MMLU-Pro and 33.45\% on MBPP, with improvements on GPQA-Diamond and LiveCodeBench as well. For Qwen3-1.7B, AutoRL improves AIME 2025 from 18.33\% to 30.00\%, HMMT 2025 from 11.67\% to 18.33\%, and Beyond AIME from 12.00\% to 12.50\%. General agent leads on AIME 2024 and IMO-AnswerBench by 1.67\% and 0.37\%, respectively. AutoRL leads on all four general-knowledge and coding benchmarks at this scale. The comparison favors specialized supervision across a broad set of tasks, with Section~\ref{sec:analysis-general-agent} further examining how its advantage develops during training.

\subsection{Flawed Recipe Fixing}
To evaluate the ability of AutoRL to fix an RL recipe with potential defects, we study a flawed setting by training LongCat-Flash-Lite on mathematical reasoning tasks with a simple grader and no KL regularization with responses of up to 131,072 tokens. Under this setting, RL training can start successfully and exhibits normal performance in the early stages, but suffers significant performance degradation in later stages. AutoRL can revise the recipe during training, while the comparison run keeps it unchanged. Both use the same evaluation protocol. Figure~\ref{fig:longcat-reasoning-eval} follows steps 0--64, with evaluation every two steps and 33 measurements per run. The AutoRL trajectory spans approximately 115.5 hours, allowing us to examine how automated diagnosis and repair affect learning as the initial defects become consequential.

\begin{figure}[!htbp]
  \centering
  \includegraphics[width=\linewidth]{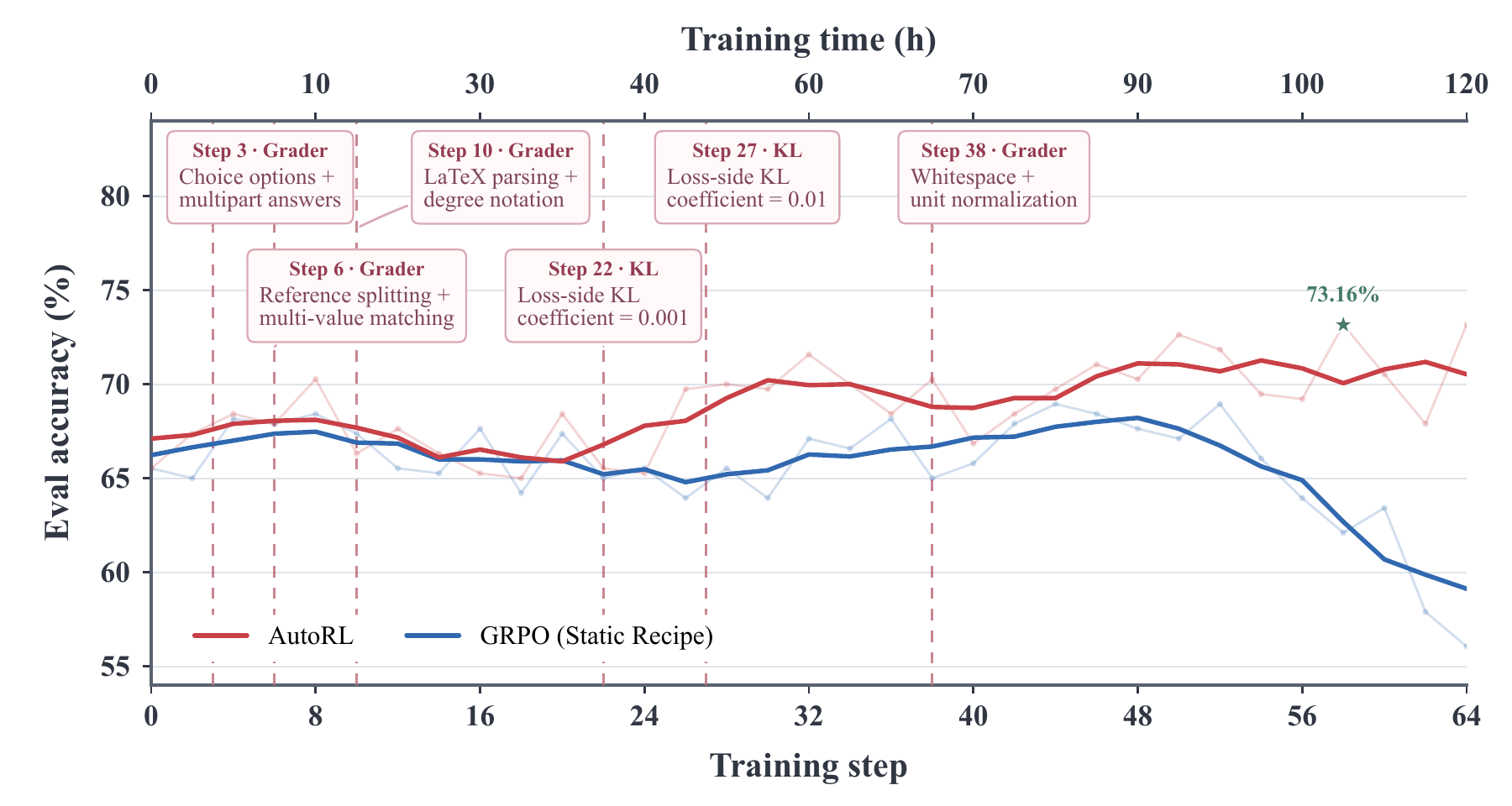}
  \caption{\textbf{Repairing the training recipe for LongCat-Flash-Lite.} AutoRL improves reasoning accuracy as the GRPO (Static Recipe) run deteriorates. Light lines and markers show raw evaluations, and dark curves show a smooth average. Dashed lines mark six representative HOT updates and no restart occurs within the experiment. The star marks the highest raw evaluation accuracy. The upper axis gives elapsed wall-clock time.}
  \label{fig:longcat-reasoning-eval}
\end{figure}

\textbf{Effectiveness of recipe repair.} As shown in Figure~\ref{fig:longcat-reasoning-eval}, the two runs initially perform similarly but diverge as training proceeds. Starting from 65.53\%, AutoRL reaches 73.16\% at step 64, improving by 7.63\%. In contrast, the GRPO (Static Recipe) run drops to 56.05\%, resulting in a gap of 17.11\%. AutoRL first reaches its peak at step 58 and matches it at step 64. The above observation illustrates how an initially workable recipe can become unproductive as learning proceeds. It also demonstrates that by monitoring the training state and performing continuous diagnosis and revision before training collapse, AutoRL can sustain effective training and improve resource utilization.

\textbf{Improving rewards and policy updates.} To understand how AutoRL revises the recipe, we examine the six interventions summarized in the figure. Four address errors in the training grader. AutoRL fixes option mapping and multipart-answer handling at step 3, followed by reference splitting and multivalue matching at step 6. It then improves LaTeX parsing and degree notation at step 10, and whitespace and unit handling at step 38. Alongside these repairs, AutoRL introduces KL regularization with a coefficient of 0.001 at step 22 and increases it to 0.01 at step 27. These modifications target complementary aspects of learning, improving the reward signal while constraining policy updates.

The two types of intervention address different sources of degradation. Grader repairs improve the learning signal, while KL regularization constrains policy drift. Applying both within the running experiment allows AutoRL to correct the recipe while preserving the model state from which learning continues. It shows that AutoRL can correct flawed recipes online from multiple perspectives, which may stem from the design of Proposers with distinct roles.

\Needspace{4\baselineskip}
\subsection{Performance Plateau Breakthrough}

Beyond correcting recipe defects, AutoRL aims to discover further improvements when a working training run reaches a plateau. To evaluate this ability, we compare AutoRL and static-recipe GRPO using Qwen3-4B, starting from the same model with the same initial training and evaluation settings. Figure~\ref{fig:qwen3-4b-plateau} shows the transition from early gains to a prolonged plateau and subsequent improvement. We examine both the performance after the plateau and the sequence of interventions leading up to it.

\begin{figure}[!htbp]
  \centering
  \includegraphics[width=\linewidth]{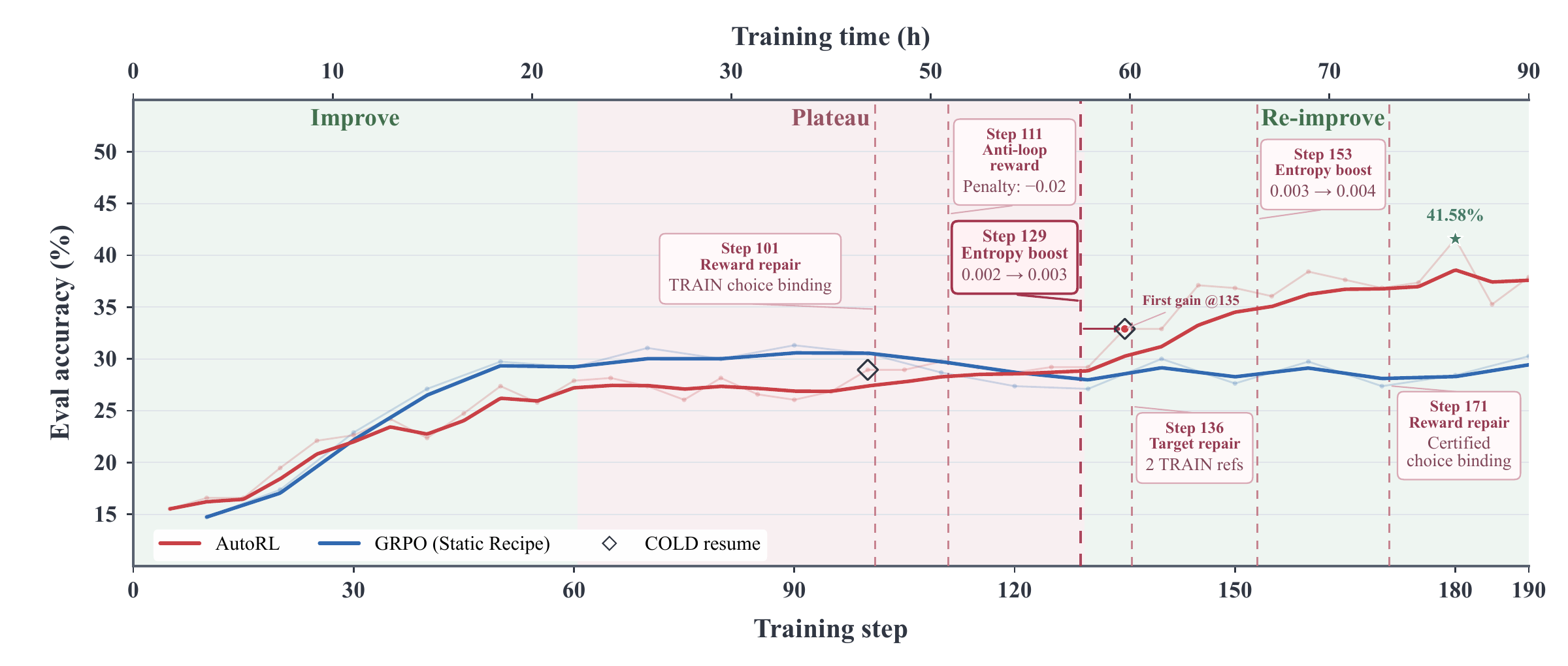}
  \caption{\textbf{Overcoming a performance plateau with Qwen3-4B.} Shaded regions indicate initial improvement, the plateau, and renewed improvement. Light lines and markers show raw evaluations, and dark curves show a smooth average. Dashed lines mark applied interventions. Hollow diamonds mark COLD resumptions from checkpoints 100 and 135, with the modifications taking effect at steps 101 and 136. The star marks the highest raw evaluation accuracy. The upper axis gives elapsed wall-clock time.}
  \label{fig:qwen3-4b-plateau}
\end{figure}

\textbf{Improvement beyond the plateau.} AutoRL reaches 41.58\% accuracy at step 180, exceeding its plateau maximum of 29.74\% by 11.84\%. This peak is also 10.26\% above the best GRPO (Static Recipe) result of 31.32\% within the plotted interval. The advantage persists at the common endpoint of step 190, where AutoRL achieves 37.89\% compared with 30.26\% for GRPO (Static Recipe). Thus, the gain is visible both at the peak and at the end of the comparison, indicating that AutoRL can sustain further improvement after an extended period of stagnation.

\textbf{Developing interventions during stagnation.} AutoRL leaves the training configuration unchanged for the first 100 steps while continuing to monitor and discuss potential improvements. During this period, it defers an entropy-adjustment proposal to gather more evidence. The first applied change fixes answer-option binding in the training grader at step 101, following a COLD resumption from checkpoint 100. AutoRL then sets the entropy coefficient to 0.001 at step 108 and adds an auxiliary penalty at step 111 for incorrect responses that satisfy a strict looping criterion. Subsequent updates raise the entropy coefficient to 0.002 and 0.003 at steps 121 and 129. The sequence moves from diagnosing the learning signal to adjusting exploration, with each decision informed by the current run.

\textbf{Connecting interventions to subsequent gains.} Accuracy remains at 29.21\% at step 130 and rises to 32.89\% at step 135, exceeding the plateau maximum for the first time. The entropy increase at step 129 is the last applied change before this rise, suggesting that stronger exploration may contribute alongside the earlier reward corrections. AutoRL continues refining the recipe afterward, repairing two training references at step 136 after resuming from checkpoint 135, increasing the entropy coefficient to 0.004 at step 153, and using a proof certificate to refine answer-option binding at step 171. It suggests that AutoRL’s ability to overcome performance plateaus and achieve further gains may stem from the cumulative benefits of ongoing interventions from multiple perspectives as training progresses.

This plateau study illustrates how AutoRL achieves additional gains through online adjustments and research over a standard normal RL recipe. AutoRL initially keeps the current recipe running while monitoring training, gathering evidence to guide targeted improvements during the subsequent plateau. After prolonged stagnation, model performance resumes improving and reaches a higher level on evaluation. It suggests that a plateau can serve as a starting point for further research, with training signals, repetitive behaviors, and exploration intensity informing testable, actionable improvements. AutoRL’s strength lies in sustaining this research process and implementing candidate modifications within ongoing training. Even after overcoming the plateau, AutoRL continues to examine reference answers and exploration settings, using feedback from one improvement to guide subsequent research. This ability to adapt its actions to the evolving training state provides an important foundation for sustained improvement over long horizons.

\subsection{Anomaly Handling in Long Training}

Automating RL supervision requires handling runtime failures and preserving the learning accumulated before an interruption. We evaluate this capability in a LongCat-Flash-Lite coding-agent run spanning 352.5 hours, approximately 14.7 days. The run adopts fully asynchronous training across $28 \times 8$ GPUs with an MoE model. Each coding trajectory involves tool use and code execution, allowing up to 400 interaction turns in the initial configuration. These long interactions, together with distributed execution and precision-sensitive MoE computations, create a demanding setting for failure diagnosis and recovery.

We study the failures that occur during this run, tracking both agent-directed repairs and restarts. Figure~\ref{fig:longcat-agentic-recovery} aligns pass@1 evaluations on SWE-related tasks with failure events and the checkpoints used to resume training. This experiment tracks how recovery decisions change as new issues emerge and whether the model continues to improve across interruptions.

\begin{figure}[!htbp]
  \centering
  \includegraphics[width=\linewidth]{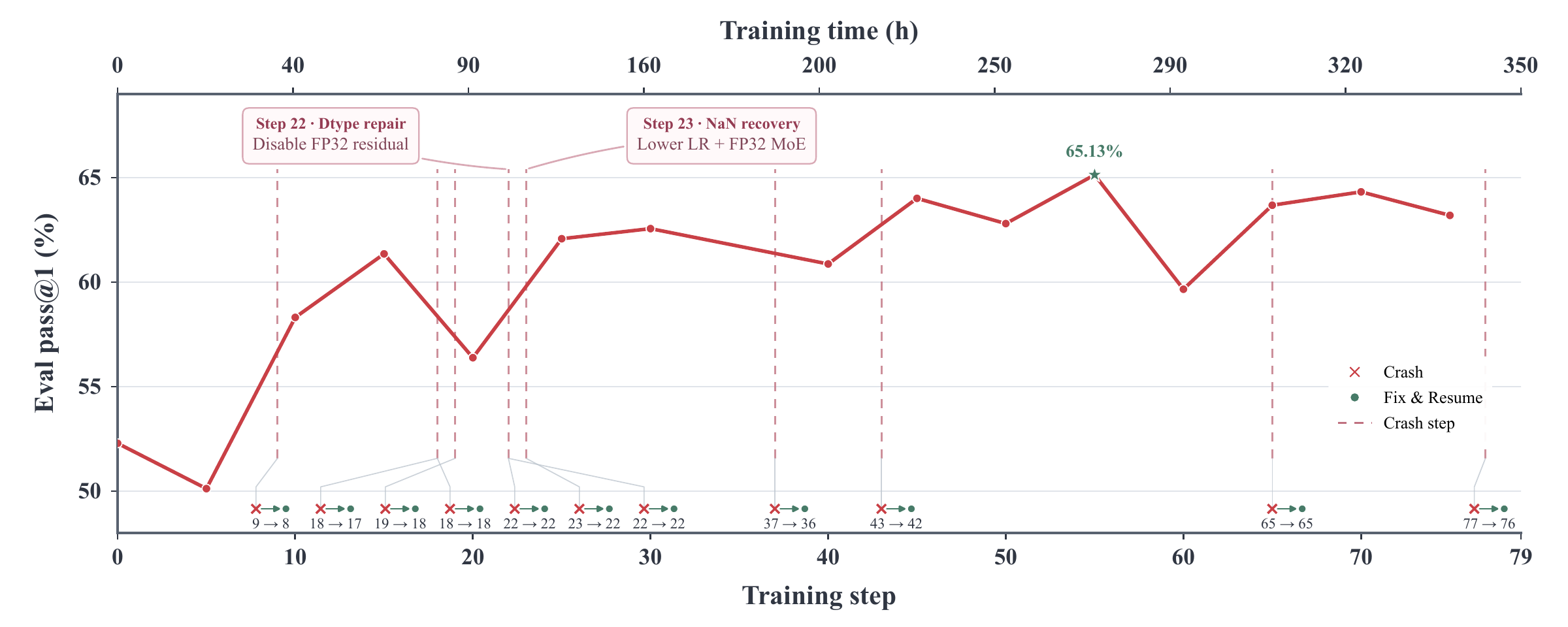}
  \caption{\textbf{Recovering from failures during LongCat-Flash-Lite coding-agent training.} The curve illustrates evaluations on SWE-related tasks, with a star marking the highest pass@1. Red dashed lines indicate failures. Paired crosses and dots connect the 11 failures to their recovery checkpoints, while the labels give the failure step and restored checkpoint. Two representative repairs are highlighted. The upper axis gives elapsed wall-clock time.}
  \label{fig:longcat-agentic-recovery}
\end{figure}

\textbf{Continued learning across failures.} The run encounters 11 distinct failures, comprising seven NaN failures, one dtype mismatch, and three worker or communication failures. Training resumes from a valid checkpoint after each failure, with agent-directed repairs and restarts together allowing the run to reach step 79. Across 15 evaluations through step 75, pass@1 rises from 52.29\% to a peak of 65.13\% at step 55 and finishes at 63.20\%. The final score remains 10.91\% above the starting model, following a peak gain of 12.84\%. Checkpoint recovery and targeted repair preserve learning progress across repeated interruptions.

\textbf{Adapting the recipe for recovery.} Repeated NaN failures and stalled jobs prompt AutoRL to revise the numerical settings before resuming from saved checkpoints. Specifically, it tightens log-ratio clipping from $[-13,5]$ to $[-5,3]$ and sets the probability-ratio spike limit to 5. It also lowers both the learning rate and minimum learning rate from $5\times10^{-6}$ to $2\times10^{-6}$, while enabling FP32 accumulation in the MoE components. By adjusting update magnitude and numerical precision, AutoRL changes the conditions that led to failure before continuing the run. Restoring model and optimizer states allows it to test the revised configuration while building on prior learning.

\textbf{Revising a repair when new problems emerge.} An attempted repair also introduces a new failure. Enabling FP32 residuals causes a dtype mismatch, which AutoRL addresses by disabling that option while retaining the reduced learning rate, MoE precision setting, and existing update constraints. Training then resumes from checkpoint 22. Several attempts reuse this checkpoint, explaining the repeated step labels in the figure. A later NaN failure near step 43 is likewise followed by resumption. This example demonstrates the importance of reassessing earlier actions, since a change intended to improve stability can introduce a different incompatibility. This recovery depends on identifying which part of a repair failed. Reverting the incompatible residual setting preserves the other numerical adjustments, allowing the next attempt to build on the earlier diagnosis.

These observations show that AutoRL can sustain stable supervision and continuous research during ongoing RL training, achieving sustained performance gains even in substantially more complex coding-agent environments involving multi-node communication, frequent numerical errors, and long-horizon interactions. These findings highlight AutoRL’s potential to reduce the supervisory burden of human RL researchers and enable long-horizon unattended RL training.

\section{Analysis}
\label{sec:analysis}

\subsection{Comparison with General Agent}
\label{sec:analysis-general-agent}

We compare AutoRL with the native general agent (Codex) on Qwen3-4B to examine how specialized supervision affects research, execution, and model quality. Both systems use \texttt{gpt-6.1-sol} as the backend LLM and start from the same model, training data, and evaluation settings. Figure~\ref{fig:qwen3-4b-general-agent-comparison} relates their research activity to the reasoning accuracy in Section~\ref{sec:experiments}. Intervention outcomes and performance use a common budget of 273 completed training steps, including steps repeated after checkpoint rollback. Proposal inventories cover the full research records, comprising first-round Proposer candidates for AutoRL and design proposals from the general agent's research notes.

\begin{figure}[!htbp]
  \centering
  \makebox[\linewidth][c]{\includegraphics[width=\linewidth]{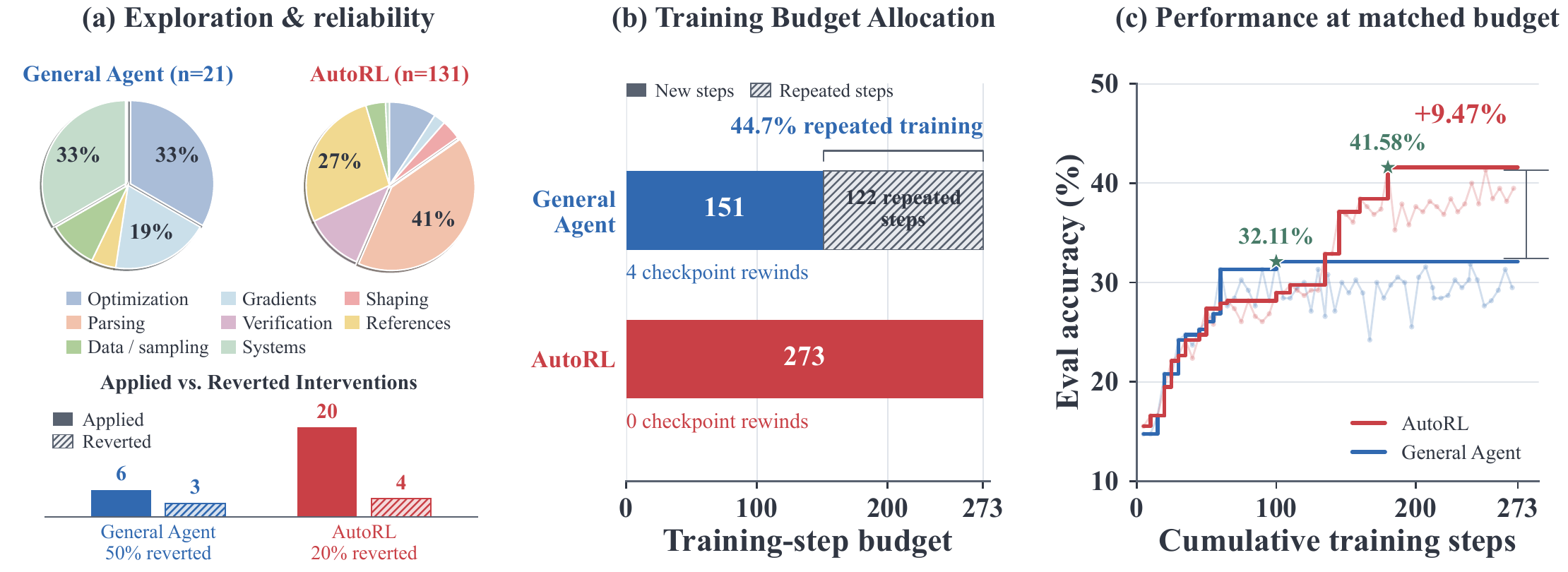}}
  \caption{\textbf{Comparison between AutoRL and general agent on Qwen3-4B.} (a) Proposal composition and applied versus reverted interventions. (b) New and repeated training steps. (c) Raw accuracy (light), best-so-far accuracy (dark), and peaks (stars).}
  \label{fig:qwen3-4b-general-agent-comparison}
\end{figure}

\textbf{Higher accuracy within the same step budget.} As shown in Figure~\ref{fig:qwen3-4b-general-agent-comparison}(c), AutoRL reaches 41.58\% accuracy, exceeding the general agent's 32.11\% by 9.47\%. The general agent reaches its best result after 100 cumulative steps, whereas AutoRL continues improving and peaks at step 180. The separation emerges after the initial gains, showing the benefit of continuing to investigate the training recipe as the run evolves. It demonstrates that AutoRL’s role specialization and RL-specific tools improve the quality of training-state perception and proposal generation, leading to better performance under the same resource budget.

\textbf{Broader investigation with a lower reversion rate.} As shown in Figure~\ref{fig:qwen3-4b-general-agent-comparison}(a), AutoRL generates 131 initial candidates across eight categories, compared with 21 general agent design proposals across five categories. The general agent emphasizes optimization and system configuration. AutoRL also investigates answer parsing, reward verification, and reference correctness, extending its search to defects in the learning signal. Within 273 training steps, AutoRL applies 20 changes and withdraws four, while the general agent applies six and withdraws three. The reversion rates are therefore 20\% and 50\%. AutoRL puts more proposals into practice while retaining a larger fraction of its applied modifications, combining a wider investigation with review before execution.

\textbf{Revising decisions while preserving training progress.} The execution paths explain another difference in how the systems use their training steps. As shown in Figure~\ref{fig:qwen3-4b-general-agent-comparison}(b), the general agent rolls back to earlier checkpoints four times, first reaching 151 distinct step indices and spending 122 of its 273 steps revisiting earlier indices (44.7\%). AutoRL reaches 273 distinct indices without checkpoint rollback. Its four withdrawn changes are reverted through online control, retaining the current model and optimizer state. Coupling research decisions with HOT control therefore lets AutoRL revise the recipe without retraining from an earlier checkpoint. This execution capability contributes to the advantage of the complete supervision framework.

\subsection{Performance with Different LLMs}
\label{sec:analysis-backbones}

\begin{figure}[H]
  \centering
  \includegraphics[width=\linewidth]{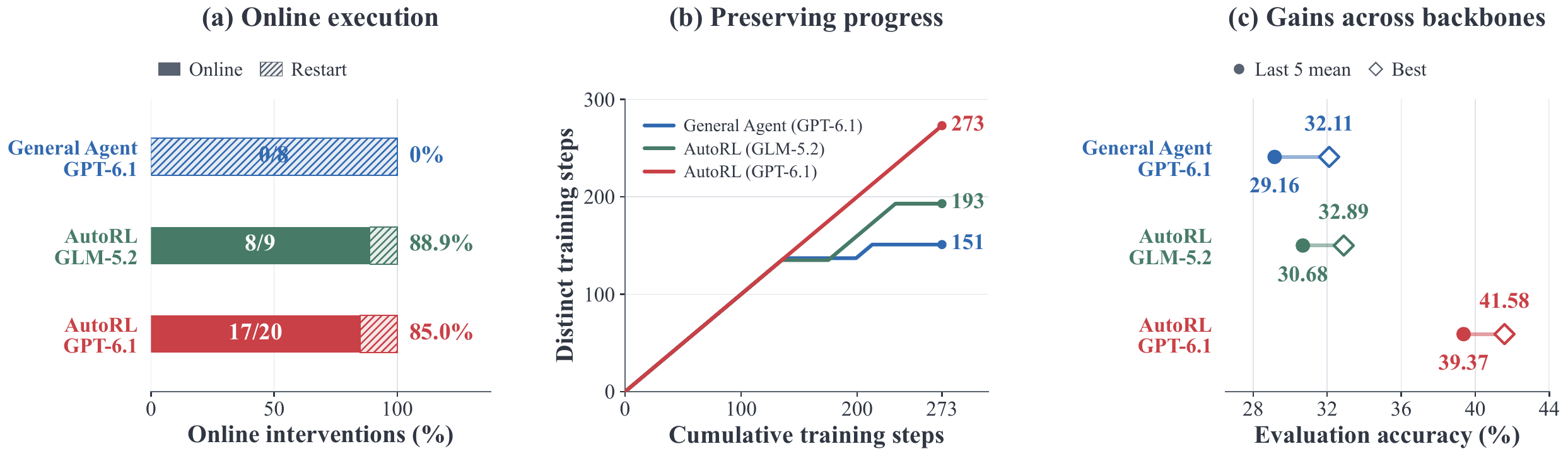}
  \caption{\textbf{AutoRL with different LLM backbones.} (a) Online intervention shares over complete runs. (b) Distinct training steps reached. (c) Best accuracy (diamonds) and the mean of the last five evaluations (circles).}
  \label{fig:qwen3-4b-backbone-system}
\end{figure}

To investigate the impact of using different LLMs as agent backends on the performance of AutoRL, we compare \texttt{gpt-6.1-sol} (GPT-6.1) and \texttt{glm-5.2} (GLM-5.2) as AutoRL's supervising LLM on the same Qwen3-4B task, retaining the GPT general agent as a reference. The runs share the initial model, training settings, and evaluation protocol. Figure~\ref{fig:qwen3-4b-backbone-system} compares progress and accuracy within 273 cumulative training steps.

\textbf{Performance varies with the supervising LLM.} Replacing GPT with GLM lowers AutoRL's peak accuracy from 41.58\% to 32.89\%, a difference of 8.69\%. Both exceed the GPT general agent's 32.11\% as shown in Figure~\ref{fig:qwen3-4b-backbone-system}(c). The mean of the last five evaluations within the shared budget is 39.37\% for GPT AutoRL and 30.68\% for GLM AutoRL, compared with 29.16\% for general agent. The advantage thus appears in both peak and late-run performance. Their research records help explain the gap between backbones. GPT investigates parsing, reward verification, and reference correctness, whereas GLM concentrates on optimization and reward shaping. The choice of LLM shapes what the system investigates and how much it improves.

\textbf{Turning different research choices into online adjustments.} Across the complete runs, 17 of GPT AutoRL's 20 interventions and eight of GLM AutoRL's nine interventions are HOT updates. The general agent implements all eight of its changes through restarting training as shown in Figure~\ref{fig:qwen3-4b-backbone-system}(a), while six fall within the 273-step comparison in Section~\ref{sec:analysis-general-agent}. Entropy regularization provides a shared adjustment to compare. GPT AutoRL applies eight entropy changes online and GLM AutoRL applies six, while the general agent implements its three changes through restarts. The framework thus gives both supervising LLMs a direct and online way to adjust the active recipe while retaining model and optimizer state.

\textbf{Using more of the budget to advance training.} Within 273 cumulative steps, GPT AutoRL reaches 273 distinct step indices and GLM AutoRL reaches 193, compared with 151 for the general agent as shown in Figure~\ref{fig:qwen3-4b-backbone-system}(b). GLM AutoRL therefore spends 42 fewer steps revisiting earlier indices. Despite its narrower research coverage, it preserves more progress and achieves higher peak and late-run accuracy than the general agent. These results illustrate complementary contributions from the LLM and the framework. The LLM determines the research ideas available to the system, while coordinated monitoring and execution help turn those ideas into useful training adjustments.

\subsection{Scope and Limitations of AutoRL}
\label{sec:analysis-scope}
To investigate the proposal scope and limitations of AutoRL, we examine what AutoRL proposes and what it adopts using the GPT-driven Qwen3-4B run from Section~\ref{sec:analysis-general-agent}. Table~\ref{tab:agent-capabilities} groups its 131 first-round candidates and 20 deployed plans by the scope of the change. Parameter tuning adjusts existing coefficients or numerical safeguards; grading and data refinement repair the learning signal; reward extensions add a shaping rule. Learning-rule redesign covers new gradient-selection or weighting rules within the training setup, providing a concrete category for innovation-oriented proposals.

\begin{table}[!htbp]
\centering
\small
\setlength{\tabcolsep}{5pt}
\renewcommand{\arraystretch}{1.12}
\caption{Research scope in the GPT-driven Qwen3-4B run. Proposed counts all proposal candidates the agents generated; adopted counts distinct plans deployed after discussion and candidate merging, including subsequently withdrawn changes.}
\label{tab:agent-capabilities}
\begin{tabularx}{\linewidth}{@{}p{0.27\linewidth}ccX@{}}
\toprule
Research category & Proposed & Adopted & Representative cases \\
\midrule
Parameter tuning & 12 & 9 & Entropy coefficients; clipping thresholds and numerical guards. \\
Grading and data refinement & 110 & 10 & Answer-choice binding; reference correction and training-pool revision. \\
Reward-mechanism extension & 5 & 1 & A −0.02 reward adjustment for incorrect looping responses. \\
Diagnostic instrumentation & 1 & 0 & Token-level logging of policy ratios and advantages. \\
Learning-rule redesign & 3 & 0 & Covariance-based gradient selection; positive-advantage and entropy-guided gradient weighting. \\
\midrule
Total & 131 & 20 & \\
\bottomrule
\end{tabularx}
\end{table}

\textbf{Most research aims to improve the existing training process.} Grading and data refinement account for 110 candidates (84.0\%), making the correctness of the learning signal the dominant research focus. The adopted plans comprise ten modifications to grading or data, nine parameter adjustments, and one reward extension. These include correcting answer-choice mismatches and erroneous references, increasing entropy regularization, and penalizing incorrect repetitive responses. AutoRL therefore carries out useful research beyond hyperparameter search while retaining the underlying GRPO learning rule.

\textbf{Learning-rule ideas rarely reach execution.} Three candidates (2.3\%) introduce a different gradient-selection or weighting rule, and none is adopted. Reviewing their discussion reveals two obstacles. The covariance-based proposal exhibits sensitivity to sequence packing and becomes inactive with constant advantages during local checks. The other two proposals reweight positive-advantage or high-entropy tokens, but the discussion favors less disruptive entropy adjustments over restarting training to introduce a new loss component. These cases locate the gap between generating an algorithmic idea and developing it into a justified training experiment. The agents explore modifications to the learning rule, yet the realized improvements come from tuning, repair, and local reward refinement.

This distribution reveals the capabilities and limitations of current agents in autonomous RL. Agents can handle recurring monitoring and refinement within a defined task, reducing the need for researchers to continuously track entropy, rewards, and runtime health. They can even analyze individual rollout data points online to repair graders, revise reference answers, and counter reward hacking. However, in our experiments, we did not observe agents within our framework generating conceptual leaps or proposing fundamental innovations that break with existing architectures. We regard this limitation as a major unresolved challenge on the path toward RSI. Nevertheless, these findings support a practical division of labor: AutoRL handles unattended supervision and incremental research, while human researchers devote more effort to conceptual breakthroughs and transformative innovations.

\section{Related Work}
\label{sec:related-work}

\Needspace{5\baselineskip}
\subsection{Autonomous Post-Training}
Automated post-training systems study training strategies, data recipes, and training tools~\citep{fang2026llmzero,he2026autosft,chen2026evotrainer}, extending to sustained multi-agent research with shared checkpoints and repeated experiments from fixed starting points~\citep{zhou2026_261021055,shi2026_260620657}. These approaches differ in their intervention scope, experiment organization, and reuse of experience. Particularly relevant are systems that incorporate feedback during training. LLMZero composes RL stages through checkpoint tree search and stops candidates using intermediate feedback. AgentJet separates mutable agent logic from persistent training servers, allowing clients to be replaced during training while preserving optimizer state; its A3R leader--worker workflow supports multi-day RL research~\citep{fang2026llmzero,fu2026_260604484}. EvoTrainer jointly evolves policies and training tools through diagnosis, branch backtesting, and reusable skills, while retaining human confirmation for training launches and version promotion in its reported setup~\citep{chen2026evotrainer}.

AutoRL organizes supervision and local research around the ongoing RL process. Whereas Athanor coordinates general research agents through shared histories and checkpoints~\citep{zhou2026_261021055}, AutoRL assigns monitoring, proposal development, and execution control to distinct roles. Evidence and intervention records connect their decisions to the active training state, while online updates and checkpoint-based resumption let them act on that state.

\Needspace{5\baselineskip}
\subsection{Autonomous Research}
Autonomous research systems organize literature, hypotheses, implementation, and experimental feedback into iterative workflows~\citep{schmidgall2025agentlaboratory,yang2025_250514738,tang2025_250518705}. Complementary benchmarks assess experimental design, execution, and iterative improvement across machine learning tasks~\citep{nathani2025_250214499,lupidi2026_260206855}. The AI Scientist-v2 combines tree search with stage management for feasibility checks, tuning, and ablation studies~\citep{yamada2025aiscientistv2}. Closer to model improvement, RSI-Master records experimental actions and evidence in a persistent experiment graph and organizes exploration through a research DAG of Workers and Reviewers, with Reviewers independently checking evidence~\citep{du2026_260935561}. These systems illustrate how research organization and review can jointly support sustained experimentation.

Experiment search also requires controlling validation cost. ImproveAnyTask checks execution using small samples of real data and short training runs before full post-training, while retaining validated strategies and scripts. AutoLLMResearch learns to select expensive configurations from low-fidelity experiments~\citep{yao2026_261006347,guo2026_260511518}. AutoRL integrates evidence review and CPU-only checks into proposal deliberation. Proposers inspect claims and test candidate logic before the Orchestrator applies an approved change to the main training process and verifies its activation. Perceptors continue observing training and trigger renewed deliberation when problems arise. This places inexpensive validation inside proposal development and uses subsequent training feedback to guide further intervention.

\Needspace{5\baselineskip}
\subsection{Multi-Agent Collaboration}
Multi-agent frameworks organize tool use and collaboration through roles, conversation protocols, and standard operating procedures~\citep{wu2023autogen,hong2023metagpt,tang2025open,tang2026beyond}. Memory systems support experience reuse through structured notes, linked retrieval, and incremental consolidation~\citep{xu2025_250212110,zhang2025_251004618}, while long-horizon evaluations examine state updates and evidence aggregation under interference~\citep{lee2026_260518565}. For collective research, Agora stores findings, hypotheses, and verification records as an append-only DAG in Git, allowing agents in separate sessions to retrieve and build on shared work~\citep{zhang2026_260918094}. AutoRL associates local histories and shared records with concrete runs, steps, interventions, and parent checkpoints, enabling agents to verify the current state and distinguish reusable experience from training history that no longer applies.

Collaboration quality also depends on candidate generation, the use of counterevidence, and final selection. Recent debate studies distinguish complementary proposal coverage from evidence-based selection, and show that unstructured discussion among homogeneous small models can increase cost without improving accuracy~\citep{zhao2026_261004686,bertalanic2026_260500914}. AutoRL organizes proposal generation, cross-examination, revision, ranking, and voting around a frozen initial observation context. Evidence gathered and CPU checks performed during deliberation are recorded and shared to support candidate revision.

\section{Conclusion}
\label{sec:conclusion}

We presented AutoRL, a multi-agent framework for automating RL supervision and research. Specialized Perceptors, complementary Proposers, and an Orchestrator connect monitoring, evidence-based deliberation, and process-level intervention, taking on work that otherwise requires repeated researcher attention. Experiments with Qwen and LongCat-Flash-Lite demonstrate recipe repair, renewed gains after plateaus, and recovery from runtime failures. On Qwen3-4B, AutoRL outperforms static recipes and a native general agent across nine benchmarks. Both GPT-6.1 and GLM-5.2 support improvements, and a coding-agent run continues for 14.7 days through 11 failures. Current agents are effective at tuning, repair, and local reward refinement within an existing recipe. Developing fundamentally different learning methods remains the harder research problem.

\bibliographystyle{unsrtnat}
\bibliography{references}

\appendix
\begingroup
\setlength{\parskip}{1.5pt}
\setlength{\textfloatsep}{10pt plus 2pt minus 2pt}
\setlength{\intextsep}{4pt plus 1pt minus 1pt}
\setlength{\floatsep}{8pt plus 2pt minus 2pt}
\makeatletter
\renewcommand{\paragraph}{%
  \@startsection{paragraph}{4}{\z@}%
    {0.8ex \@plus 0.2ex \@minus 0.1ex}{-1em}{\normalsize\bfseries}}
\makeatother
\newpage
\section{Appendix}
\label{sec:appendix}

\subsection{Task Configurations and Reproducibility}
\label{app:configurations}

Table~\ref{tab:initial-training-configurations} lists the detailed initial recipes and size of the training dataset.

\paragraph{Training data.}
Both Qwen experiments use 17,390 DAPO-Math-17K examples after removing prompts longer than 1,024 tokens. The 9,340 private LongCat-Flash-Lite mathematical reasoning problems mainly cover combinatorics and geometry, with additional problems in number theory, analysis, and probability. The coding-agent training dataset contains 700+ repository-level tasks with executable tests. Coding-agent progress is evaluated on SWE-related tasks. The Qwen evaluations additionally measure performance on general-knowledge and coding benchmarks.

\begin{table}[htbp]
\centering
\setlength{\tabcolsep}{5pt}
\caption{Initial training configurations. A rollout batch contains the listed number of prompts or tasks, each sampled with the indicated group size. Optimizer batch sizes count responses or trajectories. The token budgets for coding-agent trajectories include accumulated interaction content.}
\label{tab:initial-training-configurations}
\begin{tabular*}{\linewidth}{@{\extracolsep{\fill}}lccc@{}}
\toprule
Setting & Qwen3-1.7B / 4B & LongCat reasoning & LongCat coding-agent \\
\midrule
Training size & 17,390 & 9,340 & 700+ \\
Prompts per rollout batch & 128 & 128 & 96 \\
Trajectories per group & 8 & 8 & 16 \\
Optimizer batch size & 256 & 256 & 384 \\
Optimization epochs per batch & 1 & 1 & 1 \\
Initial learning rate & $10^{-6}$ & $10^{-6}$ & $5\times10^{-6}$ \\
Maximum prompt tokens & 1,024 & 1,024 & 1,024 \\
Generation/interaction token budget & 8,192 & 131,072 & 261,120 \\
Training sequence budget & 9,216 & 132,096 & 262,144 \\
Policy-ratio clipping range & $[0.8,1.2]$ & $[0.8,1.28]$ & $[0.8,1.26]$ \\
Initial KL coefficient & 0 & 0 & 0 \\
Training temperature / top-$p$ & 1.0 / 1.0 & 1.0 / 1.0 & 1.0 / 1.0 \\
Maximum interaction turns & 1 & 1 & 400 \\
\bottomrule
\end{tabular*}
\end{table}

\paragraph{Optimization and rollout settings.}
Qwen uses group-normalized advantages and a token-mean policy objective, while the LongCat recipes use the Dr.~GRPO advantage estimator~\citep{liu2025drgrpo}. All runs use AdamW~\citep{loshchilov2019adamw} and gradient clipping at 1.0. The AdamW parameters are $\beta_1=0.9$, $\beta_2=0.999$, and weight decay 0.1 for Qwen, and $\beta_1=0.9$, $\beta_2=0.95$, and weight decay 0.01 for LongCat. Reasoning runs start with a constant learning rate. Coding-agent training starts with both the learning rate and its minimum set to $5\times10^{-6}$ and allows rollout staleness of up to three policy versions. Entropy regularization is initially zero for Qwen, and KL regularization is initially zero for all three recipes.

\paragraph{Evaluation subsets and sampling.}
The reasoning monitoring set contains 190 problems, with 30 each from AIME 2024, AIME 2025, and HMMT 2025, and 100 from Beyond AIME. Each problem is sampled twice, yielding 380 responses per evaluation. Curves report the fraction of correct responses over this combined set. Qwen uses temperature 0.6, top-$p$ 1.0, and a 31,744-token response limit. LongCat uses temperature 1.0, top-$p$ 0.95, top-$k$ 10, and a 131,072-token response limit. Coding-agent evaluation samples one trajectory per SWE-related task at temperature 1.0 and top-$p$ 1.0.

Additional Qwen evaluations sample four responses per problem using independent random seeds, temperature 0.6, top-$p$ 1.0, and prompt/response limits of 4,096/32,768 tokens. Thinking mode is disabled in Qwen training and evaluation. MMLU-Pro uses a fixed, subject-balanced sample of 1,000 test questions (seed 42) and five in-domain demonstrations. GPQA-Diamond uses all 198 questions with zero-shot prompts. MBPP uses tasks 11-510 and up to three demonstrations. LiveCodeBench uses the cumulative release-v6 collection of 1,055 problems, with Table~\ref{tab:cross-domain-results} reporting the 350-problem Hard subset. IMO-AnswerBench uses all 400 problems in answerbench-v2. Both LiveCodeBench and IMO-AnswerBench use zero-shot prompts. Scoring uses answer equivalence for mathematics, option matching for multiple-choice questions, and executable tests for coding. Reported accuracy and pass@1 average correctness across the sampled responses.

\paragraph{Execution layout and evaluation cadence.}
Training uses Megatron-LM-based optimization~\citep{shoeybi2019megatron} and vLLM generation~\citep{kwon2023vllm} on NVIDIA H800 GPUs. Each Qwen run time-shares eight GPUs among the actor, training generator, and evaluation generator, using tensor parallelism of 2 and data parallelism of 4. LongCat reasoning initially allocates 64 GPUs: 32 for the colocated optimization roles, 16 for training generation, and 16 for evaluation generation. The fully asynchronous coding-agent recipe initially allocates 224 GPUs across 28 nodes, with 64 for optimization, 128 for training generation, and 32 for evaluation generation. GPU totals count colocated roles once. The Qwen static-recipe runs evaluate every 10 steps, while the Qwen3-4B AutoRL trajectory evaluates every 5 steps. LongCat reasoning evaluates and checkpoints every 2 steps; coding-agent training evaluates every 5 steps and checkpoints every step. Save-and-stop operations additionally produce checkpoints at intervention boundaries.

\subsection{Agent Prompts, Tools, and Memory}
\label{app:agents}

\paragraph{Agent configuration.}
The Qwen campaigns use \texttt{gpt-6.1-sol} with high reasoning effort for the Orchestrator, four Perceptors, and three Proposers. The LongCat campaigns use \texttt{glm-5.2}. The additional Qwen3-4B backbone study also uses GLM-5.2 within the AutoRL role structure. The general agent baseline runs native Codex with \texttt{gpt-6.1-sol} and all built-in tools and features enabled. Each AutoRL role uses a separate conversation. The Orchestrator coordinates seven subagents, comprising four Perceptors and three Proposers.

\paragraph{Role instructions and tool access.}
Table~\ref{tab:appendix-roles} summarizes the operational content of the role prompts. The Orchestrator first establishes the active run and its effective configuration, then schedules observations and reviews recommendations. A Perceptor completes one bounded observation cycle when awakened, records its evidence and any issue signal, and returns to waiting. Training, rollout, and evaluation observations advance with their respective metric streams; the Efficiency Perceptor can also investigate a stalled run. Proposers start from the same frozen evidence batch and contribute independently before inspecting each other's proposals. Only the Orchestrator authorizes changes to the live training process. An approved COLD change can be implemented by its designated Proposer after training has stopped and the code lock has been granted.

\begin{table}[htbp]
\centering
\small
\setlength{\tabcolsep}{4pt}
\renewcommand{\arraystretch}{1.12}
\caption{Role instructions, evidence, and outputs. Tool access follows each role's responsibility.}
\label{tab:appendix-roles}
\begin{tabularx}{\linewidth}{@{}p{0.16\linewidth}XX@{}}
\toprule
Role & Evidence and tools & Required output \\
\midrule
Orchestrator & Run status, observation receipts, recommendations; launch, stop, online control, checkpoint resumption & Decision, authorized modification, activation record, and updated run lineage \\
Train Perceptor & Loss, gradients, entropy, KL, ratios, clipping, and finite-value checks & Numerical or optimization diagnosis tied to a run and step window \\
Rollout Perceptor & TRAIN responses, rewards, truncation, group statistics, and grader inspection & Learning-signal diagnosis with representative training evidence \\
Eval Perceptor & Fixed-protocol aggregate scores and sanitized failure analysis & Performance trend, plateau, or degradation signal \\
Efficiency Perceptor & Completed steps, elapsed time, throughput, job status, and runtime logs & Progress or failure report distinguishing waiting from stalled execution \\
Tuner & Frozen evidence, current control values, source inspection, and CPU checks & Local adjustment with current and proposed values \\
Innovator & Frozen evidence, mechanism references, implementation sites, and CPU checks & Alternative mechanism hypothesis and an integration plan \\
Reviewer & Actual code path, counterexamples, prior interventions, and CPU checks & Defect analysis, repair, revision, or a recommendation to keep the current recipe \\
\bottomrule
\end{tabularx}
\end{table}

\paragraph{Observation and evidence reuse.}
Observation tools summarize metric windows using counts, means, standard deviations, slopes, and standardized deviations. Perceptors interpret these statistics alongside training behavior and attach the run, step, observation window, and source artifacts to each report. A shared barrier collects the reports before discussion begins. If no role reports an issue, training continues without a debate. Missing evidence can be requested through a targeted observation. Cached outputs retain source paths and hashes, allowing Proposers to work from the same observations without rescanning the logs.

\paragraph{Memory and session continuity.}
Each role retains recent observations and reasoning in its conversation history. Shared Markdown records summarize experiment status, data characteristics, and research findings, while a tool-accessible database stores sourced domain insights. Summaries link to the underlying observation, deliberation, control, and checkpoint records. After context compaction or session resumption, agents use these records to recover the active run and unfinished work. The launcher stores the Orchestrator's session identity and resumes interrupted sessions until the task is marked complete. Session recovery restores the supervisory workflow; checkpoint recovery restores training.

\paragraph{Evaluation visibility.}
Online supervision uses aggregate evaluation results and sanitized diagnostic reports through the evaluation observer. Proposers inspect training examples and replay stored TRAIN trajectories when developing reward or data changes. The evaluation inputs, scoring code, and sampling protocol remain fixed during intervention. The additional Qwen post-training benchmarks are run on the selected checkpoints through a separate evaluation pipeline.

\subsection{Deliberation and Low-Cost Validation}
\label{app:deliberation}

\paragraph{From independent proposals to a recommendation.}
A discussion has at most five rounds. The Proposers first diagnose the same frozen evidence independently and submit candidates. They then critique the candidates, propose combinations, and revise or withdraw their submissions. Combining proposals requires agreement from their authors. In round 4, Borda scores~\citep{zwicker2016voting} select at most two finalists. Round 5 produces a majority or unanimous recommendation and records dissent. Discussion can end after an earlier round if all three Proposers explicitly select the same eligible candidate. Leaving the recipe unchanged is also a valid recommendation.

A candidate describes the observed problem, mechanism, concrete modification, expected effects, implementation conditions, and rollback procedure. The Orchestrator then approves, rejects, defers, or requests further evidence. A request for evidence opens a targeted observation cycle, allowing the next decision to use newly collected facts while preserving the original discussion record.

\paragraph{Checks within deliberation.}
Proposers test candidates in CPU workspaces containing copies of the required source files and authorized training inputs. Each check records the candidate revision, inputs, command, tested behavior, and outputs. Successful and failed executions receive PASS and FAIL results. Timeouts and unavailable dependencies are marked INCONCLUSIVE; an inapplicable check includes a reason. Code revisions receive separate check records, and newly gathered evidence is shared at the next round boundary. These checks examine implementation behavior before deployment, while subsequent training supplies performance feedback.

\paragraph{A reward-hook revision.}
In discussion 43 of the GLM-driven Qwen3-4B run, a proposed reward hook removed the truncation penalty but accidentally bypassed the existing repetition penalty when a helper lookup failed. The Reviewer traced the fallback path, and the Tuner changed the hook to call the baseline grader before undoing only the truncation penalty. The corrected hook took effect at local training step 154.

We replayed the archived hooks after the run using four synthetic fixtures and a mock baseline grader (Table~\ref{tab:appendix-hook-check}). Starting from score 1.0, the mock grader subtracts 0.7 for repetition. The revised hook preserves this penalty in both repetition cases, returning 0.3 as intended; the original fallback incorrectly returns 1.0. Cases without repetition return 1.0 under both hooks.

\begin{table}[htbp]
\centering
\caption{CPU replay of the reward-hook fallback. The input score is 1.0 and the mock repetition penalty is 0.7. The intended change removes truncation penalties while preserving repetition penalties.}
\label{tab:appendix-hook-check}
\begin{tabular*}{\linewidth}{@{\extracolsep{\fill}}lccc@{}}
\toprule
Fixture & Initial hook & Revised hook & Expected \\
\midrule
Neither penalty & 1.0 & 1.0 & 1.0 \\
Repetition only & 1.0 & 0.3 & 0.3 \\
Truncation only & 1.0 & 1.0 & 1.0 \\
Both penalties & 1.0 & 0.3 & 0.3 \\
\bottomrule
\end{tabular*}
\end{table}

\paragraph{Complementary proposals and independent review.}
Discussion 31 combines two compatible adjustments. The Tuner proposes an entropy coefficient of 0.001, and the Innovator proposes increasing the upper clipping allowance from 0.20 to 0.28. After the Reviewer identifies that both settings belong to the same loss object, the authors merge the candidates in round 3. Both changes take effect online at local step 141. In contrast, the Orchestrator rejects discussion 1's unanimous entropy recommendation because evaluation and early training signals are still improving. For discussions 31 and 43, missing publication metadata blocked formal finalization, and the Orchestrator reviewed the recommendations directly before execution.

\subsection{Runtime Control and Checkpoint Continuity}
\label{app:runtime-control}

\paragraph{Training progress and resumption.}
Figures~\ref{fig:longcat-reasoning-eval}--\ref{fig:longcat-agentic-recovery} follow the retained checkpoint lineage. A resumed job inherits its checkpoint's step counter. Training after a rollback can therefore revisit earlier step indices. Upper axes show elapsed wall-clock time, including evaluation, waiting, and recovery. Figures~\ref{fig:qwen3-4b-general-agent-comparison} and~\ref{fig:qwen3-4b-backbone-system} instead align runs by cumulative completed steps, counting those revisits.

\paragraph{HOT controls.}
A HOT control file specifies the full set of active overrides. Supported controls include KL and advantage-estimation settings, the policy-loss object, training-grader adjustments, and save-and-stop. Setting a field to \texttt{None} restores its launch-time value, so a new control file must retain any earlier overrides that should remain active. Similarly, replacing a loss object requires carrying forward its unchanged clipping, entropy, and KL settings. The task configuration determines which controls may be used. Qwen AutoRL keeps sampling, sequence budgets, batch sizes, and learning rate fixed while investigating loss, reward, and data changes. The coding-agent task also permits learning-rate and precision changes.

The synchronous trainer loads controls at a training-step boundary, applies trainer settings, and activates training-grader changes on the rollout path. It records the effective step and source revision, retaining the previous state if activation fails. The fully asynchronous backend applies trainer-local controls before dispatching a policy update. We timestamp an applied intervention using this activation record.

\paragraph{COLD modifications and state restoration.}
Changes requiring reinitialization use COLD resumption. Before stopping, the Orchestrator decides which active HOT controls to incorporate into the next launch, reapply afterward, or remove. It requests save-and-stop, waits for shutdown, and grants the code lock for implementation. The next run restores the model, optimizer, trainer, and rollout state from a compatible checkpoint, including the saved step counter and available random and data-consumption states. If the intervention replaces training inputs, the new run checks the effective sampler and data loader. Recovery from an already failed job uses an existing complete checkpoint.

\paragraph{Lineage-aware observation.}
Each resumed run records its parent and checkpoint. Its active history follows the parent only through that checkpoint and then continues with the new run. For example, resuming checkpoint 100 retains observations through step 100; later observations from the abandoned continuation remain available as research history. The plateau study resumes checkpoints 100 and 135, with changes taking effect at steps 101 and 136. Linking controls and evaluations through these parent records lets Perceptors interpret later behavior in the context of the state actually inherited.

\subsection{Resource Accounting and Measurement Definitions}
\label{app:resource-accounting}

\paragraph{Progress, computation, and elapsed time.}
A retained step identifies progress along the active checkpoint lineage. A cumulative step counts every completed training iteration, including iterations subsequently discarded by rollback. In the Qwen recipe, each iteration collects 1,024 responses and performs four optimizer minibatch updates of size 256. The analyses in Sections~\ref{sec:analysis-general-agent} and~\ref{sec:analysis-backbones} align trajectories at 273 cumulative training steps. Wall-clock time additionally includes initialization, sampling, evaluation, waiting, checkpointing, and recovery.

A step is revisited if its numerical training index has already occurred in the campaign. The cumulative budget is therefore $B=U+R$, where $U$ counts distinct indices and $R$ counts revisits. Under this definition, AutoRL reaches 273 distinct indices and general agent reaches 151, revisiting earlier indices 122 times after four checkpoint rollbacks (Table~\ref{tab:appendix-budget}). A withdrawal of an intervention is counted separately because online control can undo a change without rewinding the checkpoint.

\begin{table}[htbp]
\centering
\caption{Qwen3-4B under a common budget of 273 completed training steps. Applied and reverted interventions use the same budget cutoff. A reversion withdraws a deployed change and can occur without rewinding a checkpoint.}
\label{tab:appendix-budget}
\begin{tabular*}{\linewidth}{@{\extracolsep{\fill}}lrr@{}}
\toprule
Measurement & General Agent & AutoRL \\
\midrule
First-visit steps & 151 & 273 \\
Revisited steps & 122 & 0 \\
Revisited share & 44.7\% & 0.0\% \\
Checkpoint rewinds & 4 & 0 \\
Applied interventions & 6 & 20 \\
Subsequently reverted interventions & 3 & 4 \\
Observed reversion rate & 50.0\% & 20.0\% \\
Highest evaluation accuracy & 32.11\% & 41.58\% \\
\bottomrule
\end{tabular*}
\end{table}

\paragraph{Proposal and intervention counts.}
The proposal inventories cover the complete research records. AutoRL has 168 discussions and 504 first-round submissions, of which 372 contain no candidate and 132 contain an action candidate. Removing one pure rollback leaves 131 candidates for new changes. General agent has 21 design proposals extracted from its notes, grouping revisions, repeated tests, and execution retries under the same proposal. Each candidate is assigned one primary category in Table~\ref{tab:appendix-proposals}, whether or not it is ultimately executed.

After discussion, AutoRL produces 51 joint action recommendations excluding NONE and pure rollback. The Orchestrator approves 28, defers 21, and rejects two. Twenty new changes reach training within 273 steps, comprising 17 HOT and three COLD interventions. Merged candidates count as one applied plan. A reversion explicitly withdraws an implemented change; a later coefficient adjustment counts as a new intervention. Reversion rates use applied and withdrawn changes observed by the same cutoff. Figure~\ref{fig:qwen3-4b-backbone-system}(a) uses the longer complete records for the backbone comparison, containing eight General Agent interventions over 343 steps, nine GLM AutoRL interventions over 478 steps, and 20 GPT AutoRL interventions over 273 steps.

\begin{table}[htbp]
\centering
\caption{Initial proposal categories over the recorded research histories. Categories describe the target mechanism; counts include proposed changes that were not executed.}
\label{tab:appendix-proposals}
\begin{tabular*}{\linewidth}{@{\extracolsep{\fill}}lrr@{}}
\toprule
Category & General Agent & AutoRL \\
\midrule
Optimization & 7 & 12 \\
Gradient design & 4 & 3 \\
Reward shaping & 0 & 5 \\
Answer parsing & 0 & 54 \\
Reward verification & 0 & 15 \\
Reference repair & 1 & 36 \\
Data and sampling & 2 & 5 \\
Systems & 7 & 1 \\
\midrule
Total & 21 & 131 \\
Categories covered & 5 & 8 \\
\bottomrule
\end{tabular*}
\end{table}

\paragraph{Evaluation and visualization.}
All numerical comparisons use raw evaluations. Figure~\ref{fig:longcat-reasoning-eval} overlays a centered five-point moving average, while Figure~\ref{fig:qwen3-4b-plateau} uses the same step-adjusted piecewise exponential smoothing for both trajectories. Peak markers refer to raw scores. Figure~\ref{fig:longcat-agentic-recovery} connects 15 evaluations after removing duplicate recovery evaluations. Upper time axes interpolate observed step/time anchors and round labels to the nearest 10 hours. Figure~\ref{fig:autorl-teaser} uses smoothed reasoning trajectories with illustrative intervention markers.

\subsection{Intervention, Recovery, and Failure Case Studies}
\label{app:case-studies}

\paragraph{Repairing the LongCat reasoning recipe.}
Table~\ref{tab:appendix-recipe-events} lists the six selected HOT interventions in Figure~\ref{fig:longcat-reasoning-eval}. Four repair the training grader and two adjust KL regularization. The model continues training through all six without restarting.

\begin{table}[htbp]
\centering
\setlength{\tabcolsep}{4pt}
\caption{Selected LongCat reasoning interventions. Steps indicate runtime activation. All grader changes apply to TRAIN; the evaluation grader remains fixed.}
\label{tab:appendix-recipe-events}
\begin{tabularx}{\linewidth}{@{}rX@{}}
\toprule
Step & Applied change \\
\midrule
3 & Repair multiple-choice option mapping and multipart-answer extraction. \\
6 & Split compound references and compare multivalue answers as multisets. \\
10 & Improve LaTeX and array parsing and normalize degree notation. \\
22 & Introduce KL regularization with coefficient 0.001. \\
27 & Increase the KL coefficient to 0.01. \\
38 & Normalize whitespace and handle units for numeric-only references. \\
\bottomrule
\end{tabularx}
\end{table}

\paragraph{Developing improvements during the Qwen plateau.}
Table~\ref{tab:appendix-plateau-events} lists selected changes around the Qwen3-4B plateau. Controls remain unchanged through step 100. The first intervention repairs answer-option binding after resuming checkpoint 100, followed by entropy, looping-penalty, and reference updates. The entropy change at step 129 precedes the first evaluation above the plateau maximum at step 135; the two-reference repair takes effect at step 136.

\begin{table}[htbp]
\centering
\setlength{\tabcolsep}{4pt}
\caption{Selected Qwen3-4B interventions around the plateau. COLD changes resume from the immediately preceding checkpoint shown in the text; HOT changes retain the running state.}
\label{tab:appendix-plateau-events}
\begin{tabularx}{\linewidth}{@{}rlX@{}}
\toprule
Step & Mode & Applied change \\
\midrule
101 & COLD & Repair TRAIN answer-option binding after checkpoint 100. \\
108 & HOT & Set the entropy coefficient to 0.001. \\
111 & HOT & Apply an auxiliary reward of $-0.02$ to already incorrect responses satisfying a strict looping criterion. \\
121 & HOT & Increase the entropy coefficient to 0.002. \\
129 & HOT & Increase the entropy coefficient to 0.003. \\
136 & COLD & Repair two TRAIN references after checkpoint 135. \\
153 & HOT & Increase the entropy coefficient to 0.004. \\
171 & HOT & Refine TRAIN answer-option binding using a verified finite-orbit proof certificate. \\
\bottomrule
\end{tabularx}
\end{table}

\paragraph{Eleven failures and their recovery paths.}
Table~\ref{tab:appendix-recovery-events} matches the 11 failures in the 352.5-hour LongCat coding-agent campaign to checkpoint-loading receipts. Crash steps denote the last completed update, or the parent checkpoint if none completed, rather than the internal update index raising the exception. Recovery latency spans the error timestamp to re-entry into the learning loop, including intermediate restart attempts.

\begin{table}[htbp]
\centering
\small
\setlength{\tabcolsep}{3pt}
\caption{Failure and recovery log for LongCat coding-agent training. Times are UTC+8 in 2026. The arrow gives completed step $\rightarrow$ restored checkpoint. Latency ends at the resumption shown in the figure.}
\label{tab:appendix-recovery-events}
\begin{tabularx}{\linewidth}{@{}lllr rX@{}}
\toprule
ID & Failure time & Symptom & Step & Minutes & Recovery path \\
\midrule
F1 & 09-24 21:05 & NaN & $9\!\to\!8$ & 38.9 & Checkpoint restart \\
F2 & 09-26 12:12 & NaN & $18\!\to\!17$ & 39.1 & Checkpoint restart \\
F3 & 09-27 00:14 & NaN & $19\!\to\!18$ & 59.1 & Agent: new job \\
F4 & 09-27 06:10 & NaN & $18\!\to\!18$ & 36.6 & Checkpoint restart \\
F5 & 09-28 07:11 & NaN & $22\!\to\!22$ & 295.1 & Agent: ratio protection \\
F6 & 09-28 18:18 & NaN & $23\!\to\!22$ & 106.5 & Agent: LR/precision \\
F7 & 09-29 00:15 & Dtype mismatch & $22\!\to\!22$ & 101.2 & Agent: revert residual \\
F8 & 10-01 15:51 & Worker exit & $37\!\to\!36$ & 40.6 & Checkpoint restart \\
F9 & 10-02 18:37 & NaN & $43\!\to\!42$ & 46.4 & Checkpoint restart \\
F10 & 10-06 14:42 & Communication & $65\!\to\!65$ & 38.1 & Checkpoint restart \\
F11 & 10-08 02:55 & Communication & $77\!\to\!76$ & 37.1 & Checkpoint restart \\
\bottomrule
\end{tabularx}
\end{table}

Recovery from the 11 failures combines checkpoint resumption with targeted repairs, including repeated attempts for some failures. F3 resumes checkpoint 18 in a replacement job. F5 resumes checkpoint 22 after tightening log-ratio clipping from $[-13,5]$ to $[-5,3]$ and setting the ratio spike limit to 5. F6 lowers the learning rate and its minimum from $5\times10^{-6}$ to $2\times10^{-6}$ and enables FP32 accumulation during MoE unpermutation and FP32 residuals. The residual change causes F7's dtype mismatch and is withdrawn, retaining the other controls before resuming checkpoint 22. F9 recovers from checkpoint 42.

\paragraph{Data adaptation during continued training.}
The coding-agent task pool changes from an initial pool of 700+ tasks to 656 in epoch 2 and 453 in epoch 3. Epoch 4 restores 58 previously removed tasks and removes 55 others, producing 456 tasks. Selection therefore revisits the full pool rather than only pruning the preceding subset. The final executed selection contains 271 tasks. These changes adjust the composition of the supplied data without introducing a new training objective.

\par
\endgroup

\clearpage
\ifPDFTeX\end{CJK*}\fi
\end{document}